\documentclass[letterpaper, 10 pt, journal, twoside]{ieeetran}  

\IEEEoverridecommandlockouts                              

\IEEEpubid{\begin{minipage}{\textwidth}\ \\[12pt] 
  \copyright 2020 IEEE. Personal use of this material is permitted.  Permission from IEEE must be obtained for all other uses.
\end{minipage}}

\usepackage{times}
\usepackage[pdftex]{graphicx}
\usepackage{subfigure}
\usepackage{amsmath,amssymb,amsopn,amstext,amsfonts}
\usepackage{cancel}
\usepackage[space]{cite}
\usepackage{pdfsync}
\usepackage{balance}
\usepackage{color}
\usepackage{mathtools}
\usepackage{bm}

\usepackage{diagbox}
\usepackage{float}
\usepackage{epstopdf}
\usepackage{pifont}
\usepackage{fixltx2e}
\usepackage{amsmath}
\usepackage{multirow}
\usepackage{url}
\usepackage{verbatim}
\usepackage[linkcolor=black,citecolor=black,urlcolor=black,colorlinks=true]{hyperref}
\usepackage{soul,xcolor}
\usepackage[linesnumbered,ruled,vlined]{algorithm2e}
\usepackage{subfiles}
\graphicspath{{./img_compress/}{../img_compress/}}
\DeclareGraphicsExtensions{.png,.jpg,.eps,.pdf,.jpeg}

\title{
  FUEL: Fast UAV Exploration using Incremental Frontier Structure and Hierarchical Planning
}

\author{Boyu Zhou, Yichen Zhang, Xinyi Chen and Shaojie Shen%
\thanks{Manuscript received: October, 15, 2020; Accepted December, 13, 2020.
This paper was recommended for publication by Editor Pauline Pounds upon
evaluation of the Associate Editor and Reviewers' comments. 
This work was supported by Research Grants Council (RGC) project no.16213717, ITC project no.ITT/027/19GP, HDJI lab.
All authors are with the Department of Electronic and Computer Engineering, Hong Kong University of Science and Technology, Hong Kong, China. {\tt\footnotesize $\{$bzhouai, yzhangec, xchencq, eeshaojie$\}$@connect.ust.hk}.
Digital Object Identifier (DOI): see top of this page.
}%
}

\begin{document}
\maketitle


\begin{abstract}
  Autonomous exploration is a fundamental problem for various applications of unmanned aerial vehicles(UAVs).
  Existing methods, however, were demonstrated to insufficient exploration rate, due to the lack of efficient global coverage, conservative motion plans and low decision frequencies.  
  In this paper, we propose \textbf{FUEL}, a hierarchical framework that can support \textit{F}ast \textit{U}AV \textit{E}xp\textit{L}oration in complex unknown environments.
  We maintain crucial information in the entire space required by exploration planning by a frontier information structure (FIS), which can be updated incrementally when the space is explored.
  Supported by the FIS, a hierarchical planner plans exploration motions in three steps, which find efficient global coverage paths, refine a local set of viewpoints and generate minimum-time trajectories in sequence. 
  We present extensive benchmark and real-world tests, in which our method completes the exploration tasks with unprecedented efficiency (3-8 times faster) compared to state-of-the-art approaches.
  Our method will be made open source to benefit the community\footnote{To be released at \url{https://github.com/HKUST-Aerial-Robotics/FUEL}}.
    
\end{abstract}


\section{Introduction}
\label{sec:intro}


\IEEEPARstart{U}{nmanned} aerial vehicles, especially quadrotors have gained widespread popularity in various applications, such as inspection, precision agriculture, and search and rescue.
Among the tasks, autonomous exploration, in which the vehicle explores and maps the unknown environment to gather information, is a fundamental component. 

Various exploration planning methods have been proposed in recent years, with some real-world experiments presented \cite{cieslewski2017rapid,schmid2020efficient,meng2017two,selin2019efficient}.  
However, most of them demonstrate a low/medium exploration rate, which is unsatisfactory for many large-scale real-world applications.
First of all, many existing planners plan exploring motions in greedy manners, such as maximizing the immediate information gain, or navigating to the closest unknown region.
The greedy strategies ignore global optimality and therefore result in low overall efficiency. 
Besides, most methods generate rather conservative motions in order to guarantee information gain and safety simultaneously in previously unknown environments.
Low-speed exploration, however, disallows quadrotors to fully exploit their dynamic capability to fulfill the missions.
Last but not least, many methods suffer from high computational overhead and can not respond quickly and frequently to environmental changes.
However, to enable faster exploration, it is desirable to replan new motions immediately whenever new information of the environment is available.

\begin{figure}[t]
   \begin{center}          
      \subfigure[A cluttered environment for the exploration tests.]
      {\includegraphics[width=0.75\columnwidth]{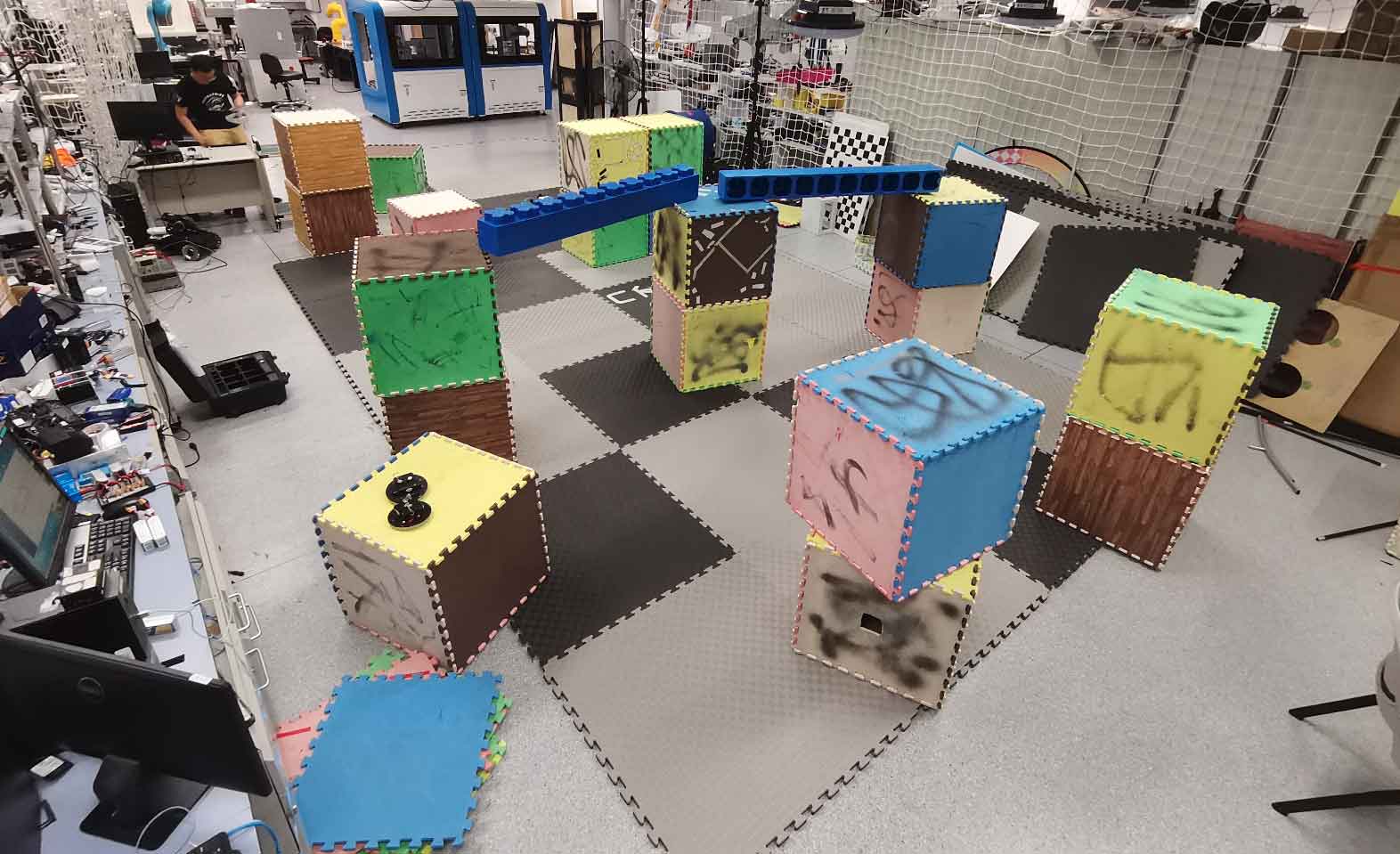} \label{fig:intro1}}
      \subfigure[The online built map and executed trajectory.]
      {\includegraphics[width=0.75\columnwidth]{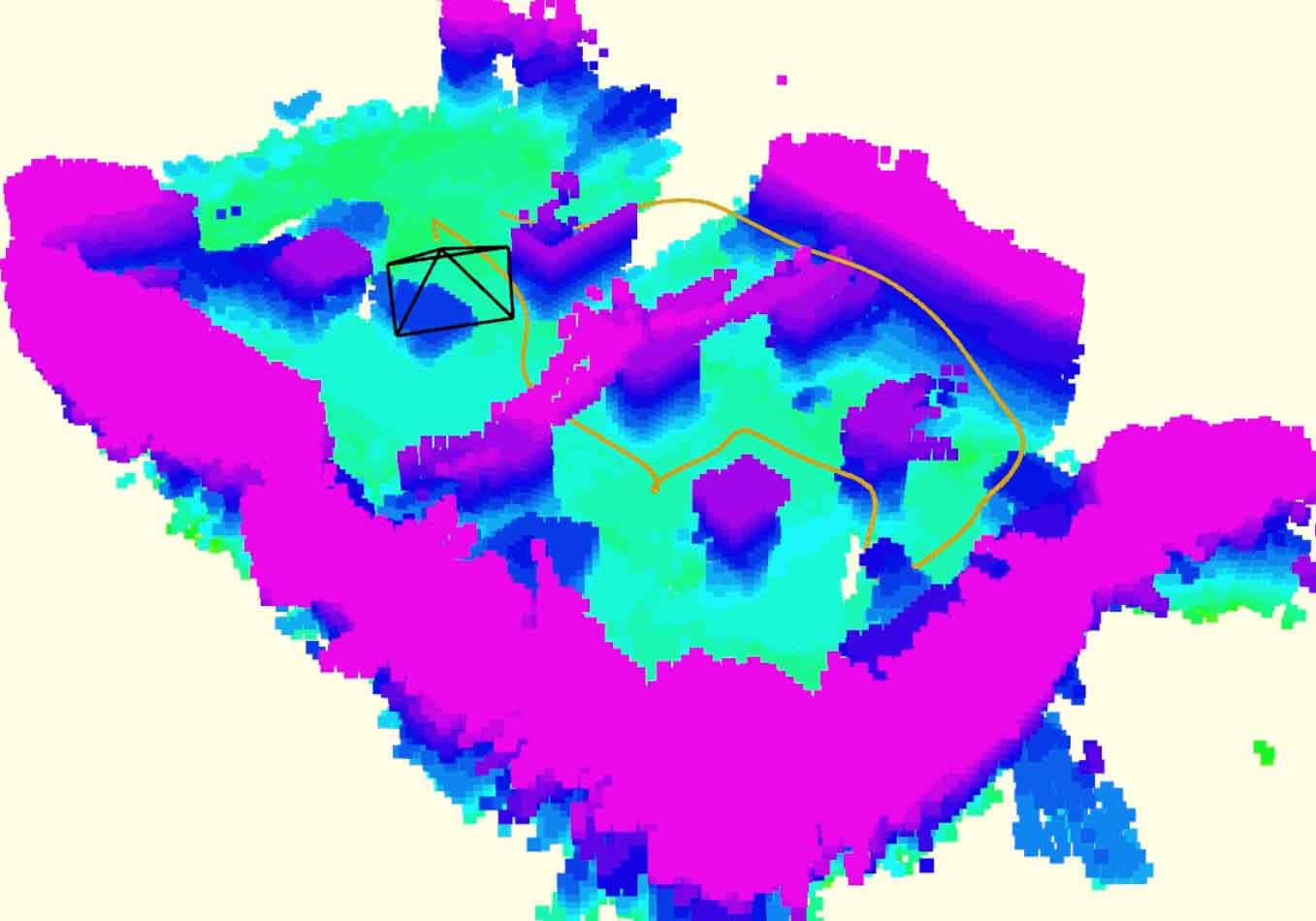} \label{fig:intro2}}
      \vspace{-0.4cm}
	\end{center}
   \caption{\label{fig:intro} A quadrotor autonomous exploration test conducted in a complex indoor scene. Video of the experiments is available at: \url{https://www.youtube.com/watch?v=_dGgZUrWk-8}.
   }
   \vspace{-0.7cm}
\end{figure} 

Motivated by the above facts, this paper proposes \textbf{FUEL}, a hierarchical framework that can support \textbf{F}ast \textbf{U}AV \textbf{E}xp\textbf{L}oration in complex environments.
We introduce a frontier information structure (FIS), which contains essential information in the entire space required by exploration planning.
The structure can be updated efficiently and incrementally when new information is collected, so it is capable of supporting high-frequency planning.
Based on the FIS, a hierarchical planner generates exploring motion in three coarse-to-fine steps. 
It starts by finding a global exploration tour that is optimal in the context of accumulated environment information.
Then, viewpoints on a local segment of the tour are refined, further improving the exploration rate.
Finally, a safe, dynamically feasible and minimum-time trajectory is generated.
The planner produces not only efficient global coverage paths, but also safe and agile local maneuvers.
Moreover, the planner is triggered whenever unvisited regions are explored, so that the quadrotor always responds promptly to any environmental changes, leading to consistently fast exploration.

We compare our method with classic and state-of-the-art methods in simulation.
Results show that in all cases our method achieves complete exploration in much shorter time (3-8 times faster).
What's more, we conduct fully onboard real-world exploration in various challenging environments.
Both the simulation and real-world tests demonstrate an unprecedented performance of our method compared to state-of-the-art ones.
To benefit the community, we will make the source code public.
The contributions of this paper are summarized as follows:

1) An incrementally updated FIS, which captures essential information of the entire explored space and facilitates exploration planning in high frequency.

2) A hierarchical planning method, which generates efficient global coverage paths, and safe and agile local maneuvers for high-speed exploration.

3) Extensive simulation and real-word tests that validate the proposed method. The source code of our system will be made public.

\section{Related Work}
\label{sec:related}

\subsection{Exploration Path Planning}
\label{subs:related_exploration}

Robotic exploration, which uses mobile robots to map unknown environments, has been studied for years.
Some of the works focus on exploring the space quickly\cite{cieslewski2017rapid, dharmadhikari2020motion}, as this paper does. 
Meanwhile, other methods place more emphasis on accurate reconstruction\cite{schmid2020efficient, song2017online}. 
Among the various proposed methods, the frontier-based approaches are one type of classic approaches.
The methods are first introduced in \cite{yamauchi1997frontier} and evaluated more comprehensively afterwards in \cite{julia2012comparison}.
To detect frontiers in 3D space, a stochastic differential equation-based method is proposed in \cite{shen2012stochastic}.
In the original method\cite{yamauchi1997frontier}, the closest frontier is selected as the next target.
\cite{cieslewski2017rapid} presented a different scheme. 
In every decision, it selects the frontier within the FOV that minimizes the velocity change to maintain a consistently high flight speed. 
This scheme is shown to outperform the classic method\cite{yamauchi1997frontier}.
In \cite{deng2020robotic}, a differentiable measure of information gain based on frontiers is introduced, allowing paths to be optimized with gradient information.

Sampling-based exploration, as another type of major approaches, generate viewpoints randomly to explore the space.
These methods are closely related to the concept of next best view (NBV) \cite{connolly1985determination}, which computes covering views repeatedly to obtain a complete model of a scene. 
\cite{bircher2016receding} first used NBV in 3D exploration, in which it expands RRTs with accessible space and executes the edge with the highest information gain in a receding horizon fashion.
The method was extended to consider uncertainty of localization\cite{papachristos2017uncertainty}, visual importance of different objects \cite{dang2018visual} and inspection tasks \cite{bircher2018receding} later.
To avoid discarding the expanded trees directly, roadmaps are constructed in \cite{witting2018history, wang2019efficient} to reuse previous knowledge. 
\cite{schmid2020efficient} maintains and refines a single tree continuously using a rewiring scheme inspired by RRT*.
To achieve faster flight, \cite{dharmadhikari2020motion} samples safe and dynamically feasible motion primitives directly and execute the most informative one.

There are also methods combining the advantages of frontier-based and sampling-based approaches.
\cite{charrow2015information,selin2019efficient} plan global paths toward frontiers and sample paths locally.
\cite{charrow2015information} also presented a gradient-based method to optimize the local path.
\cite{meng2017two} samples viewpoints around frontiers and finds the global shortest tour passing through them.
\cite{song2017online} generates inspection paths that cover the frontier completely using a sampling-based algorithm.

Most of existing methods make decision greedily and does not account for the dynamics of the quadrotor, which leads to inefficient global tours and conservative maneuvers.
In contrast, we plan tours that efficiently cover the entire environment and generate dynamically feasible minimum-time trajectories to enable agile flight.

\subsection{Quadrotor Trajectory Planning}
\label{subs:related_trajectory}

Trajectory planning for quadrotor has been widely studied, which can be categorized into the hard-constrained and soft-constrained approaches in major.
The former is pioneered by minimum-snap trajectory\cite{MelKum1105}, whose closed-form solution was presented \cite{RicBryRoy1312} later.
Based on \cite{MelKum1105}, \cite{CheLiuShe2016, fei2018icra, ding2019efficient} extract convex safe regions for safe trajectory generation. 
To obtain a more reasonable time allocation, fast marching\cite{fei2018icra}, kinodynamic search\cite{ding2019efficient} and mixed integer-based methods \cite{tordesillas2019faster} are proposed.
\cite{fei2018icra} also introduced an efficient B\'ezier curve-based method to guarantee feasibility.


Soft-constrained methods typically formulate a non-linear optimization trading off several objectives.
Recently \cite{oleynikova2016continuous, fei2017iros, usenko2017real, zhou2019robust, zhou2020robust, zhou2020raptor} applied them to local replanning, demonstrating their attractive performance.
The methods were revived by \cite{ratliff2009chomp} and extended to continuous-time trajectories\cite{oleynikova2016continuous} later.
To relieve the issue of local minima, \cite{fei2017iros} initializes the optimization with collision-free paths.
\cite{usenko2017real} introduces uniform B-splines for replanning.
More recently, \cite{zhou2019robust} further exploited B-splines and demonstrated fast flight in field tests.
\cite{zhou2019robust} is further improved with topological guiding paths and perception-awareness in \cite{zhou2020robust, zhou2020raptor}.

In this paper, we base our trajectory planning on \cite{zhou2019robust} but extend it to optimize all parameters of B-splines. 
In this way, the total trajectory time can be minimized so that the unknown space is explored with a higher navigation speed.

\begin{figure}[t]
	\begin{center}          
		{\includegraphics[width=0.9\columnwidth]{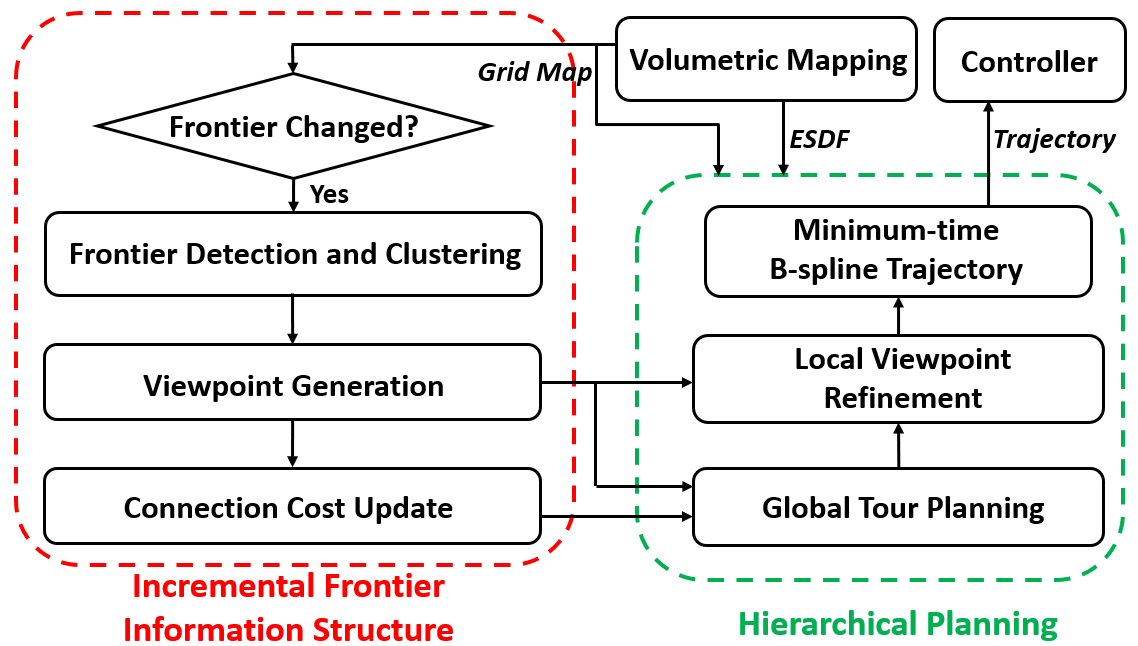}}       
      \vspace{-0.5cm}
	\end{center}
   \caption{\label{fig:overview} An overview of the proposed exploration framework.
   }
   \vspace{-0.9cm}
\end{figure} 

\section{System Overview}
\label{sec:overview}

The proposed framework operates upon a voxel grid map. 
As illustrated in Fig.\ref{fig:overview}, it is composed of an incremental update of the FIS (Sect.\ref{sec:frontier}) and a hierarchical exploration planning approach (Sect.\ref{sec:planning}). 
Whenever the map is updated using sensor measurements, it is examined whether any frontier clusters are influenced.
If so, FISs of influenced clusters are removed while new frontiers along with their FISs are extracted (Sect.\ref{sec:frontier}).
After that, the exploration planning is triggered, which finds global exploration tour, refines local viewpoints, and generates trajectory to a selected viewpoint successively (Sect.\ref{sec:planning}). 
The exploration is considered finished if no frontier exists.

\section{Incremental Frontier Information Structure}
\label{sec:frontier}

As presented in classic frontier-based exploration \cite{yamauchi1997frontier}, frontiers are defined as known-free voxels right adjacent to unknown voxels, which are grouped into clusters to guide the navigation.
Traditionally, the extracted information is too coarse to do fine-grained decision making.
Besides, frontiers are retrieved by processing the entire map, which is not scalable for large scenes and high planning frequencies.
In this work, we extract richer information from frontiers to enable more elaborate planning, and develop an incremental approach to detect frontiers within the locally updated map.  

\begin{figure}[t]
	\begin{center}          
    {\includegraphics[width=0.99\columnwidth]{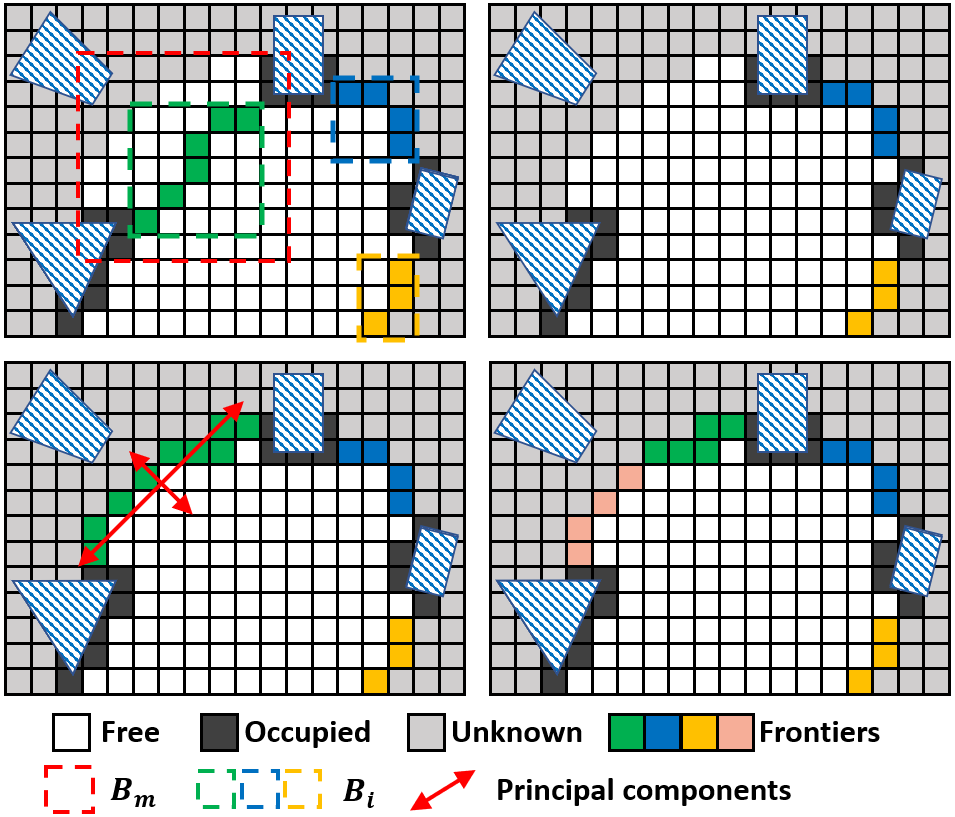}}       
      \vspace{-0.8cm}
  \end{center}
   \caption{\label{fig:frontier_detection} Incremental frontier detection and clustering.
   Top: detecting and removing outdated frontiers. 
   Bottom: new frontier is detected (left) and PCA is performed, the large cluster is split into two smaller ones (right).
   }
   \vspace{-0.8cm}
\end{figure} 

\subsection{Frontier Information Structure}
\label{subs:frontier_structure}

A frontier information structure $ FI_i $ is computed when a new frontier cluster $ F_i $ is created.
It stores all cells $C_{i}$ belonging to the cluster and the average position $ \mathbf{p}_{\text{avg},i} $ of $ C_{i} $.
The axis-aligned bounding box (AABB) $ B_{i} $ of the cluster is also computed, in order to accelerate the detection of frontier changes (Sect.\ref{subs:frontier_detection}).
To serve the exploration planning (Sect.\ref{sec:planning}), candidate viewpoints $\textit{VP}_i$ are generated around the cluster. 
Besides, a doubly linked list $ L_{\text{cost},i} $ containing connection costs between $ F_i $ and all other clusters is computed.
Data stored by a FIS is listed in Tab.\ref{tab:structure}.

\subsection{Incremental Frontier Detection and Clustering}
\label{subs:frontier_detection}

As depicted in Fig.\ref{fig:frontier_detection}, every time the map is updated by sensor measurements, the AABB of the updated region $ B_m $ is also recorded, within which outdated frontier clusters are removed and new ones are searched.
It starts with going through all clusters and returning only those whose AABBs ($B_i$) intersect with $ B_m $.
Then, precise checks are conducted for the returned clusters, among which the ones containing cells that are no longer frontier are removed.
These two processes are inspired by the broad/narrow phase collision detection algorithms\cite{ericson2004real}, which eliminate most unaffected clusters in a fast way and significantly reduce the number of expensive precise checks.

After the removal, new frontiers are searched and clustered into groups by the region growing algorithm, similar to the classic frontier-based method.
Among the groups, the ones with a small number of cells typically resulting from noisy sensor observations are ignored. 
The remaining groups, however, may contain large-size clusters which are not conducive to distinguishing distinctive unknown regions and making elaborate decisions.
Therefore, we perform Principal Component Analysis (PCA) for each cluster and split it into two uniform ones along the first principal axis, if the largest eigenvalue exceeds a threshold.
The split is conducted recursively so that all large clusters are divided into small ones.

\begin{figure}[t]
	\begin{center}          
		{\includegraphics[width=0.9\columnwidth]{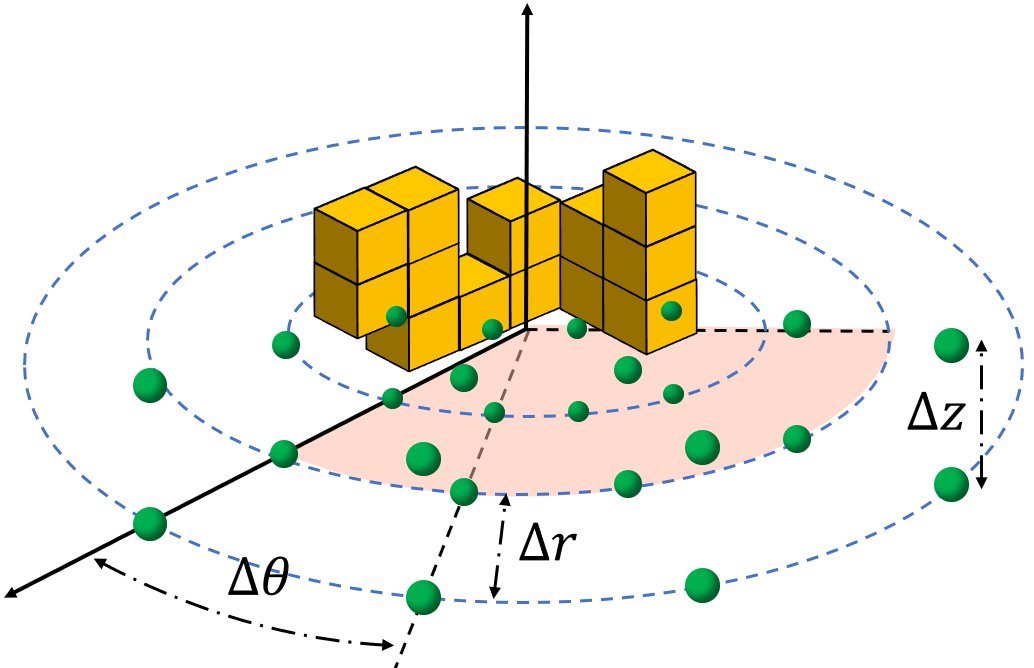}}       
      \vspace{-0.5cm}
	\end{center}
   \caption{\label{fig:viewpoint} Generating candidate viewpoints for a frontier cluster.
   Within the cylindrical coordinate system centered at the average position of the cluster, points are sampled uniformly.
   }
\end{figure} 

\begin{table}[t]
  \centering
  \caption{\label{tab:structure} Data contained by a FIS $ FI_i $ of cluster $F_i$.}
  \begin{tabular}{cc} 
  \hline\hline
   \textbf{Data}    &  \textbf{Explanation}    \\
  \hline
    $C_i$    &  Frontier cells that belong to the cluster    \\
    $ \mathbf{p}_{\text{avg},i} $    &  Average position of $C_i$  \\
    $ B_{i} $    &  Axis-aligned bounding box of $C_i$  \\
    $\textit{VP}_{i}$    &  Viewpoints covering the cluster  \\
    $ L_{\text{cost},i} $   &  Doubly linked list of connection costs to all other clusters  \\
  \hline\hline
  \end{tabular}
  \vspace{-1.0cm}
\end{table}

\subsection{Viewpoint Generation and Cost Update}
\label{subs:frontier_viewpoint}

Intuitively, a frontier cluster implies a potential destination to explore the space. 
However, unlike previous approaches which simply navigate to the center of a cluster, we desire more elaborate decision making.
To this end, when a cluster $F_i$ is created, we generate a rich set of viewpoints $\textit{VP}_i = \left\{ \mathbf{x}_{i,1}, \mathbf{x}_{i,2}, \cdots, \mathbf{x}_{i,n_i} \right\} $ covering it, where $ \mathbf{x}_{i,j} = \left( \mathbf{p}_{i,j}, \xi_{i,j} \right) $.
$\textit{VP}_i$ are found by uniformly sampling points in the cylindrical coordinate system whose origin locates at the cluster's center, as displayed in Fig.\ref{fig:viewpoint}.
For each of the sampled points $ \mathbf{p} $ lying within the free space, the yaw angle $ \xi $ is determined as the one maximizing sensor coverage to the cluster, by using a yaw optimization method similar to\cite{witting2018history}.
The coverage is evaluated as the number of frontier cells that comply with the sensor model and are not occluded by occupied voxels.
Then, viewpoints with coverage higher than a threshold are reserved and sorted in descending order of coverage.
We reserve at most $ N_{\text{view}} $ viewpoints in $\textit{VP}_i$ ($n_i \le N_{\text{view}} $) to make the local viewpoint refinement (Sect.\ref{sec:planning}) tractable.

To perform global planning of exploration tour (Sect.\ref{sec:planning}), a connection cost between each pair of clusters $ (F_{k_1}, F_{k_2}) $ is required.
Let $ t_{\text{lb}}(\mathbf{x}_{k_1,j_1}, \mathbf{x}_{k_2,j_2}) $ denotes a time lower bound when moving between two viewpoints $ \mathbf{x}_{k_1,j_1} $ and $ \mathbf{x}_{k_2,j_2} $, it is computed by:
\begin{align}
  \label{equ:time_bound}  
  \hspace{-0.2cm} t_{\text{lb}}&(\mathbf{x}_{k_1,j_1}, \mathbf{x}_{k_2,j_1}) = \max \bigg\{ \frac{\text{length}\left(P \left(  \mathbf{p}_{k_1,j_1},  \mathbf{p}_{k_2,j_2} \right)\right)}{v_{\text{max}}},  \\ \nonumber
  & \frac{\min\left( |\xi_{k_1,j_1}-\xi_{k_2,j_2}|, 2\pi-|\xi_{k_1,j_1}-\xi_{k_2,j_2}| \right)}{\dot{\xi}_{\text{max}}} \bigg\}
  \end{align}
where $ P \left(  \mathbf{p}_{k_1,j_1},  \mathbf{p}_{k_2,j_2} \right) $ denote a collision-free path between $ \mathbf{p}_{k_1,j_1} $ and $ \mathbf{p}_{k_2,j_2} $ found by a path searching algorithm, $ v_{\text{max}} $ and $ \dot{\xi}_{\text{max}} $ are the limits of velocity and angular rate of yaw.
For each pair $ (F_{k_1}, F_{k_2}) $, we select the viewpoints with highest coverage and estimate the cost as $ t_{\text{lb}}(\mathbf{x}_{k_1,1}, \mathbf{x}_{k_2,1}) $, in which $ P \left(  \mathbf{p}_{k_1,1},  \mathbf{p}_{k_2,1} \right) $ is searched on the voxel grid map using the A* algorithm.

Note that computing connection costs between all pairs of $ N_{\text{cls}} $ clusters from scratch requires $ O(N_{\text{cls}}^2) $ A* searching, which is considerably expensive.
Fortunately, the costs can also be computed in an incremental manner.
When any outdated clusters are removed (Sect.\ref{subs:frontier_detection}), associated cost items in $ L_{\text{cost},i} $ of all remaining FISs are erased. 
After that, connection costs from each new cluster to all other clusters are computed and inserted into $ L_{\text{cost},i} $.
Suppose there are $ k_{\text{new}} $ new clusters in each frontier detection, the above update scheme takes $ O(k_{\text{new}} \cdot N_{\text{cls}} ) $ time.
Practically, $ k_{\text{new}} $ is small and can be regarded as a constant factor, resulting in a linear time update of connection costs.

\section{Hierarchical Exploration Planning}
\label{sec:planning}

\begin{figure}[t]
	\begin{center}          
		{\includegraphics[width=0.85\columnwidth]{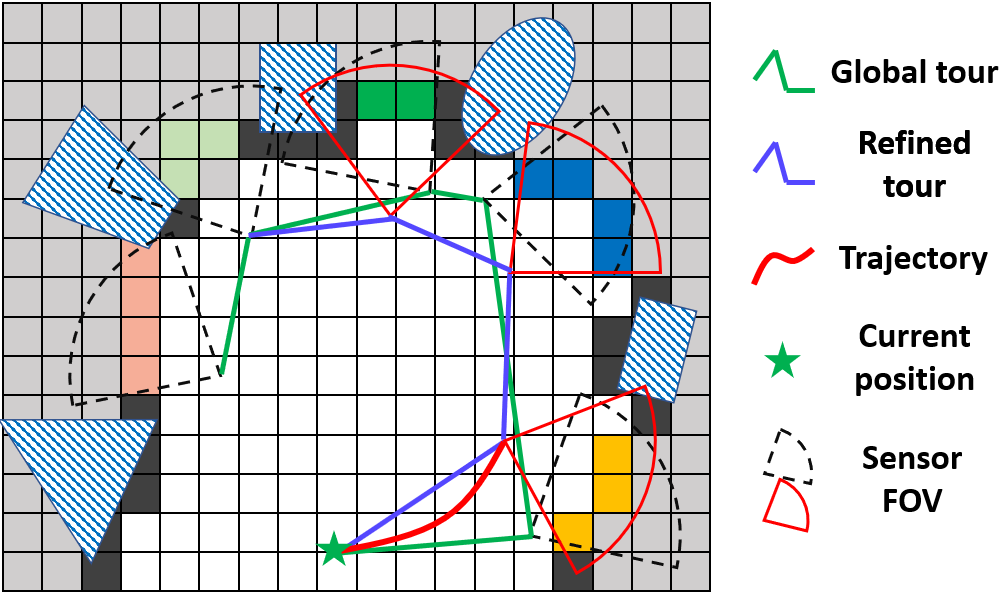}}       
      \vspace{-0.5cm}
	\end{center}
   \caption{\label{fig:planning} Generating exploration motion in three coarse-to-fine steps:
   (1) the shortest tour covering frontier clusters in the entire environment is found.
   (2) a local segment of the global tour is refined.
   (3) a safe minimum-time trajectory is generated to the first viewpoint on the refined tour.
   }
   \vspace{-0.9cm}
\end{figure}

Instead of adopting greedy exploration strategies or generating conservative maneuvers, we produce global paths to cover the frontiers efficiently and plan safe and agile motions for faster flight.
Our planner takes inspiration from the recent hierarchical quadrotor planning paradigm \cite{CheLiuShe2016, fei2018icra, zhou2019robust}, and makes decisions in three phases, as shown in Fig.\ref{fig:planning}.

\subsection{Global Exploration Tour Planning}
\label{subs:planning_global}

Our exploration planning begins with finding a global tour to cover existent frontier clusters efficiently. 
Inspired by\cite{meng2017two}, we formulate it as a variant of the Traveling Salesman Problem (TSP), which computes an open-loop tour starting from the current viewpoint and passing viewpoints at all clusters. 
We reduce this variant to a standard Asymmetric TSP (ATSP) that can be solved quickly by available algorithms, by properly designing the engaged cost matrix $ \mathbf{M}_{\text{tsp}} $.

Assume there are $ N_{\text{cls}} $ clusters totally, $ \mathbf{M}_{\text{tsp}} $ corresponds to a $ N_{\text{cls}}+1$ dimensions square matrix.
The major part is the $ N_{\text{cls}} \times N_{\text{cls}} $ block composed of the connection cost between each pair of frontier clusters, which is computed as:
\begin{align}
  & \mathbf{M}_{\text{tsp}}(k_1,k_2) = \mathbf{M}_{\text{tsp}}(k_2,k_1) \\ \nonumber
  &= t_{\text{lb}}(\mathbf{x}_{k_1,1}, \mathbf{x}_{k_2,1}), \ k_1, k_2 \in \left\{1,2,\cdots,N_{\text{cls}} \right\} 
\end{align}
As mentioned in Sect.\ref{subs:frontier_viewpoint}, this information is maintained when frontiers are detected. 
Thus, the $ N_{\text{cls}} \times N_{\text{cls}} $ block can be filled without extra overhead.

The first row and column of $\mathbf{M}_{\text{tsp}}$ are associated with the current viewpoint $\mathbf{x}_0 = \left( \mathbf{p}_0, \xi_0 \right)$ and $N_{\text{cls}}$ clusters.
Starting from $\mathbf{x}_0$, the cost to the $k$-th cluster is evaluated by:
\begin{align}
  \label{equ:start_connect_cost}
  \hspace{-0.5cm} \mathbf{M}_{\text{tsp}}(0,k) = t_{\text{lb}}(\mathbf{x}_0,\mathbf{x}_{k,1}) + w_{\text{c}} \cdot c_{\text{c}}(\mathbf{x}_{k,1}),\\ \nonumber
   k \in \left\{1,2, \cdots, N_{\text{cls}} \right\}
\end{align}
here a motion consistency cost $ c_{\text{c}}(\mathbf{x}_{k,1}) $ is introduced, which is generally computed as:
\begin{equation}
   c_{\text{c}}(\mathbf{x}_{k,j}) = \cos^{-1} \frac{(\mathbf{p}_{k,j}-\mathbf{p}_0)\cdot \mathbf{v}_0}{\left\| \mathbf{p}_{k,j}-\mathbf{p}_0 \right\| \left\|\mathbf{v}_0 \right\|} 
\end{equation}
where $\mathbf{v}_0$ is the current velocity.
In some cases, multiple tours have comparable time lower bound, so back-and-forth maneuvers may be generated in successive planning steps and slow down the progress.
We eliminate this inconsistency with $ c_{\text{c}}(\mathbf{x}_{k,1}) $, which penalizes large changes in flight direction.

Our problem is different from standard TSP whose solution is a closed-loop tour.
However, we can reduce it to an ATSP by assigning zero connection costs from other clusters to the current viewpoint:
\begin{equation}
   \mathbf{M}_{\text{tsp}}(k,0) = 0, \ k \in \left\{0, 1,2, \cdots, N_{\text{cls}} \right\}
\end{equation}
In this way, going back to the current viewpoint in any closed-loop tours contributes no extra cost, so each closed-loop tour always contains an open-loop one with an identical cost. 
As a result, we can obtain the optimal open-loop tour by finding the optimal closed-loop one and retrieving its equal-cost open-loop tour.

\begin{figure}[t]
	\begin{center}          
		{\includegraphics[width=0.8\columnwidth]{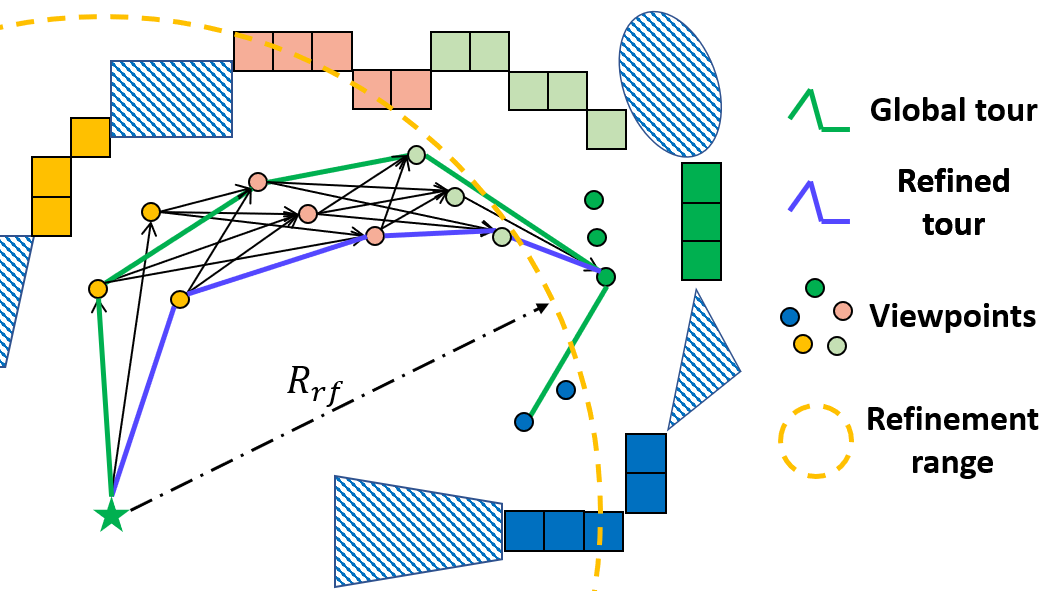}}       
      \vspace{-0.5cm}
	\end{center}
   \caption{\label{fig:planning_local} Refining viewpoints locally using the graph search approach.
   Along a truncated segment of the global tour, multiple viewpoints of each visited cluster are considered to select the optimal set of viewpoints.
   }
   \vspace{-0.5cm}
\end{figure} 

\subsection{Local Viewpoint Refinement}
\label{subs:planning_local}

The global tour planning finds a promising order to visit all clusters.
Nonetheless, it involves only a single viewpoint of each cluster, which are not necessarily the best combination among all viewpoints.

To this end, a richer set of viewpoints on a truncated segment of the global tour are considered to further improve the exploration rate, by using a graph search approach, as depicted in Fig.\ref{fig:planning_local}.
We found consecutive clusters whose viewpoints on the global tour are closer than $ R_{\text{rf}} $ to the current position.
To simplify the notation, suppose that $ F_i, 1 \le i \le N_{\text{rf}} $ are the considered clusters. 
We create graph nodes for their viewpoints $ \textit{VP}_i $ and the current viewpoint $ \mathbf{x}_0 $.
Then each node is connected to other nodes associated with the next cluster with a directed edge, which compose a directed acyclic graph capturing possible variation of the truncated tour.
We utilize the Dijkstra algorithm to search for the optimal local tour $ \Xi = \left\{ \mathbf{x}_{1,j_1}, \mathbf{x}_{2,j_2}, \cdots, \mathbf{x}_{N_{\text{rf}},j_{N_{\text{rf}}}} \right\} $ that minimizes the cost: 
\begin{align}
\label{equ:cost_local}
   c_{\text{rf}}(\Xi) &= t_{\text{lb}}(\mathbf{x}_0,\mathbf{x}_{1,j_1}) + w_{\text{c}} \cdot c_{\text{c}}( \mathbf{x}_{1,j_1} ) \\ \nonumber
   &+ t_{\text{lb}}(\mathbf{x}_{N_{\text{rf}},j_{N_{\text{rf}}}},\mathbf{x}_{N_{\text{rf}}+1,1}) + \sum_{k=1}^{N_{\text{rf}}-1} t_{\text{lb}}(\mathbf{x}_{k,j_k},\mathbf{x}_{k+1,j_{k+1}})
\end{align}
which also consists of time lower bounds and motion consistency.
Note that it is straight forward to incorporate information gain\cite{bircher2016receding,selin2019efficient} to Equ.\ref{equ:cost_local}, however, evaluating information gain for numerous viewpoints is expensive.
Practically, we find that simply adopting viewpoints based on their coverages is much faster and leads to consistently satisfactory results.

\subsection{Minimum-time B-spline Trajectory}
\label{subs:planning_traj}

Given the discrete viewpoints, continuous trajectories are required for smooth navigation.
Our quadrotor trajectory planning is based on a method\cite{zhou2019robust} that generates smooth, safe and dynamically feasible B-spline trajectories.
We go one step further to optimize all parameters of B-splines, so that the total trajectory time is minimized to enable the quadrotor to fully utilize its dynamic capability.

As the quadrotor dynamics are differentially flat\cite{MelKum1105}, we plan trajectories for the flat outputs $ \mathbf{x} \in \left( x,y,z,\xi \right) $.
Let $ \mathbf{X}_{\text{cb}} = \left\{ \mathbf{x}_{\text{c},0}, \mathbf{x}_{\text{c},1}, \cdots, \mathbf{x}_{\text{c},N_b} \right\} $ where $ \mathbf{x}_{\text{c},i} = \left( \mathbf{p}_{\text{c},i}, \xi_{\text{c},i} \right) $ be the $ N_b+1 $ control points of a $ p_b $ degree uniform B-spline, and $ \Delta t_{b} $ be the knot span.
We find the B-spline that trades-off smoothness and total trajectory time, and satisfies safety, dynamic feasibility and boundary state constraints.
It can be formulated as an following optimization problem:
\begin{equation}
   \underset{ \mathbf{X}_{\text{c},b}, \Delta t_b }{\arg \min } \ f_{\text{s}}+ w_{\text{t}} T + \lambda_{\text{c}} f_{c}+ \lambda_{\text{d}}\left( f_{\text{v}} + f_{\text{a}} \right) + \lambda_{\text{bs}} f_{\text{bs}}
\end{equation}
Similar to \cite{zhou2019robust}, $ f_{\text{s}} $ is the elastic band smoothness cost:
\begin{equation}
   f_{\text{s}} = \sum_{i = 0}^{N_b-2} \mathbf{s}_{i}^{\text{T}} \mathbf{R}_{\text{s}} \mathbf{s}_{i}, \ \mathbf{s}_{i} = \mathbf{x}_{\text{c},i+2}-2\mathbf{x}_{\text{c},i+1}+\mathbf{x}_{\text{c},i}
\end{equation}
in which $ \mathbf{R}_{\text{s}} $ is the penalty matrix:
\begin{equation}
   \mathbf{R}_{\text{s}} = \begin{bmatrix}
      w_{\text{s},\text{p}}\mathbf{I}_3 & \mathbf{0} \\
      \mathbf{0}^{\text{T}} & w_{\text{s},\xi}
      \end{bmatrix}
\end{equation}
$ T $ is the total trajectory time depending on $ \Delta t_b $ and the number of B-spline segments:
\begin{equation}
   T = (N_{b}+1-p_b) \cdot \Delta t_b
\end{equation}
$ f_{\text{c}} $, $ f_{\text{v}} $ and $ f_{\text{a}} $ are the penalties to ensure safety and dynamic feasibility.
Given the following function:
\begin{equation}
	\mathcal{P}(\tau_1, \tau_2) = \left\{
	\begin{array}{cl}
	(\tau_1-\tau_2)^{2} & \ \  \tau_1 \le \tau_2 \\
	0 & \ \ \text{else}
	\end{array}
	 \right.
\end{equation}
$ f_{\text{c}} $ is evaluated as:
\begin{equation}
\label{equ:collision_cost}
   f_{\text{c}} = \sum\limits_{i=0}^{N_{b}} \ \mathcal{P}\left( d\left(\mathbf{p}_{\text{c},i}\right), d_{\text{min}} \right)
\end{equation}
where $ d(\mathbf{p}_{\text{c},i}) $ is the distance of point $\mathbf{p}_{\text{c},i}$ to the nearest obstacle, which can be obtained from the Euclidean signed distance field (ESDF) maintained by the mapping module.
Practically a clearance larger than our quadrotor's radius (typically $d_{\min} \ge$ 0.5 m) ensures safety in complex scenes.
$ f_{\text{v}} $ and $ f_{\text{a}} $ penalize infeasible velocity and acceleration:
\begin{equation}
\label{equ:vel_cost}
f_{\text{v}} = \sum_{i=0}^{N_b-1} \left\{\sum\limits_{\mu \in \{x,y,z\}} \mathcal{P}(v_{\text{max}}, |\dot{p}_{\text{c},i,\mu}|) + \mathcal{P}(\dot{\xi}_{\text{max}}, |\dot{\xi}_{\text{c},i}|) \right\} 
\end{equation}
\begin{equation}
\label{equ:acc_cost}
f_{\text{a}} = \sum_{i=0}^{N_b-2} \left\{\sum\limits_{\mu \in \{x,y,z\}} \mathcal{P}(a_{\text{max}}, |\ddot{p}_{\text{c},i,\mu}|) + \mathcal{P}(\ddot{\xi}_{\text{max}}, |\ddot{\xi}_{\text{c},i}|) \right\} 
\end{equation}
in which the control points of derivatives are utilized:
\begin{align}
    & \dot{\mathbf{x}}_{\text{c},i} = \left[ \dot{p}_{\text{c},i,x}, \dot{p}_{\text{c},i,y}, \dot{p}_{\text{c},i,z}, \dot{\xi}_{\text{c},i} \right]^\text{T} =  \frac{\mathbf{x}_{\text{c},i+1} - \mathbf{x}_{\text{c},i}}{\Delta t_b} \\
    & \ddot{\mathbf{x}}_{\text{c},i} = \left[ \ddot{p}_{\text{c},i,x}, \ddot{p}_{\text{c},i,y}, \ddot{p}_{\text{c},i,z}, \ddot{\xi}_{\text{c},i} \right]^\text{T} =  \frac{\mathbf{x}_{\text{c},i+2} - 2\mathbf{x}_{\text{c},i+1} + \mathbf{x}_{\text{c}, i}}{\Delta t_{b}^{2}}
\end{align}
In Equ.\ref{equ:collision_cost}, \ref{equ:vel_cost} and \ref{equ:acc_cost}, the convex hull property of B-spline is utilized to ensure the feasibility efficiently.
For brevity we refer the reader to \cite{zhou2019robust} for more details.

Lastly, we set the 0th to 2nd order derivatives at the start to the instantaneous state $ ( \mathbf{x}_0, \dot{\mathbf{x}}_0, \ddot{\mathbf{x}}_0) $ for smooth motion. 
The 0-th order derivative at the end is also determined by the viewpoint $ \mathbf{x}_{\text{next}} $ to be visited.
In implementation we use cubic B-splines, so the associated cost is: 
\begin{align}
\label{equ:cost_boundary}
& f_{\text{bs}} = \left\| \frac{\mathbf{x}_{\text{c},0}+4\mathbf{x}_{\text{c},1}+\mathbf{x}_{\text{c},2}}{6} - \mathbf{x}_{0} \right \|^2 + \left\| \frac{ \dot{\mathbf{x}}_{\text{c},0}+ \dot{\mathbf{x}}_{\text{c},1}}{2} - \dot{\mathbf{x}}_0 \right\|^2 \\ \nonumber
   &+ \left\| \ddot{\mathbf{x}}_{\text{c},0} - \ddot{\mathbf{x}}_0 \right\|^2 + \left\| \frac{\mathbf{x}_{\text{c},N_b-2}+4\mathbf{x}_{\text{c},N_b-1}+\mathbf{x}_{\text{c},N_b}}{6} - \mathbf{x}_{\text{next}} \right \|^2
\end{align}

\begin{figure}[t]
	\begin{center}          
		{\includegraphics[width=0.85\columnwidth]{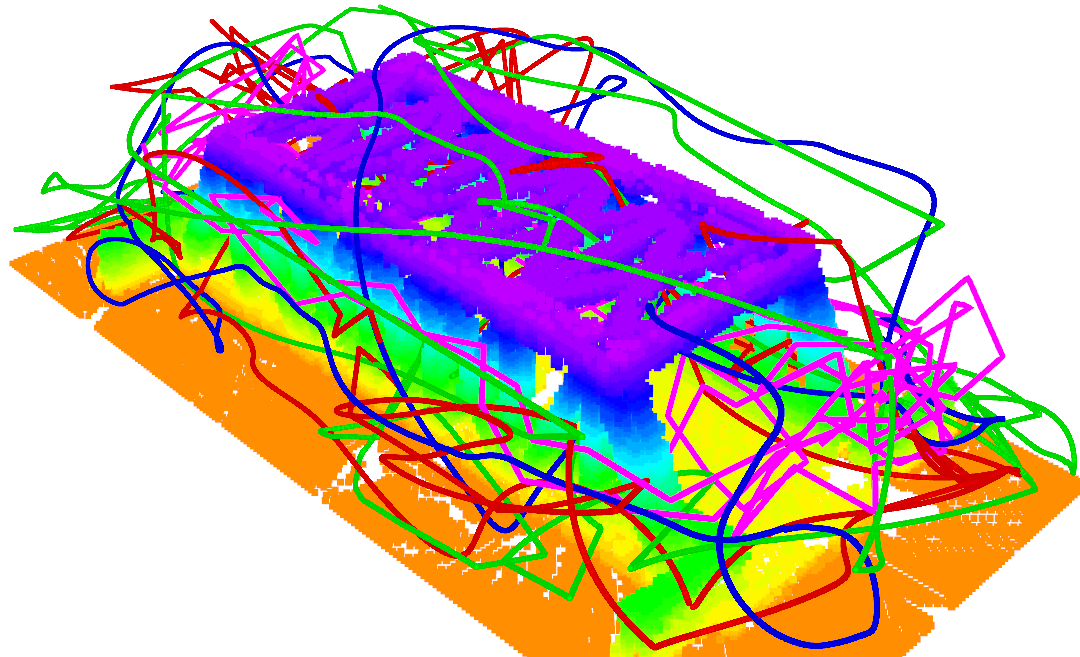}}       
      \vspace{-0.6cm}
	\end{center}
   \caption{\label{fig:bridge_compare} Benchmark comparison of the proposed method, classic frontier method\cite{yamauchi1997frontier}, rapid frontier method\cite{cieslewski2017rapid} and NBVP\cite{bircher2016receding} in a 3D space containing a bridge.
   The overall exploration paths are shown as blue, red, green and pink respectively.
   }
   \vspace{-0.3cm}
\end{figure} 

\begin{table}
   \centering
   \caption{\label{tab:benchmark} Exploration statistic in the bridge and large maze scenarios.}
   \label{tab:real1}
   \begin{tabular}{p{0.5cm}cp{0.35cm}p{0.35cm}p{0.35cm}p{0.35cm}p{0.35cm}p{0.35cm}p{0.35cm}p{0.35cm}} 
   \hline\hline
   \textbf{Scene}           & \textbf{Method}               & \multicolumn{4}{c}{\textbf{Exploration time (s)} }            & \multicolumn{4}{l}{\textbf{Flight distance (m)}}               \\ 
   \cline{3-10}
                               & \multicolumn{1}{c}{}          & \textbf{Avg} & \textbf{Std} & \textbf{Max} & \textbf{Min} & \textbf{Avg} & \textbf{Std} & \textbf{Max} & \textbf{Min}  \\
   \multirow{4}{*}{Bridge}     & Classic\cite{yamauchi1997frontier}  & 575 & 53 & 643 & 511 & 250 & 42 & 285 & 190          \\
                               & Rapid\cite{cieslewski2017rapid}     & 288 & 15 & 305 & 264 & 286 & 13 & 303 & 269               \\
                               & NBVP\cite{bircher2016receding}      &  857 & 117 & 1018 & 740 & 322 & 47 & 377 & 261           \\
                               & Proposed                            &          104 & 1.5 & 105 & 102 & 165 & 3.8 & 170 & 161             \\
   \hline
   \multirow{4}{*}{\begin{tabular}[c]{@{}c@{}}Large \\Maze\end{tabular}} & Classic\cite{yamauchi1997frontier}                     & 814& 104& 961& 721         & 419 & 63 & 509 & 373           \\
                               & Rapid\cite{cieslewski2017rapid}                                                                  & 669 & 68 & 766 & 613 & 469 & 32 & 514 & 440               \\
                               & NBVP\cite{bircher2016receding}                                                                   & 1037 & 152 & 1253 & 925 & 1539 & 262 & 1898 & 1279              \\
                               & Proposed                                                                                         & 168 & 16 & 192 & 156 & 280 & 20 & 310 & 264            \\ 
   \hline\hline
   \end{tabular}
   \vspace{-1.3cm}
\end{table}

\section{Results}
\label{sec:results}

\subsection{Implementation Details}
\label{subs:results_detail}
We set $ w_{\text{c}} = 1.5 $ in Equ.\ref{equ:start_connect_cost} and \ref{equ:cost_local}.
In global tour planning, the ATSP is solved using a Lin-Kernighan-Helsgaun heuristic solver \cite{helsgaun2000effective}.
In local viewpoint refinement, we reserve at most $ N_{\text{view}} = 15 $ viewpoints in each FIS and truncate the global tour within radius $ R_{\text{rf}}= 5.0 $.
For trajectory optimization, we use $ w_{\text{s,p}}=5.0 $, $ w_{\text{s},\xi}=2.5 $, $ w_{\text{t}}=1.0 $, $ \lambda_{\text{c}} = \lambda_{\text{bs}} = 10.0 $, $\lambda_{\text{d}} = 2.0$ and $ d_{\min} = 0.4 $ and solve the problem with a general non-linear optimization solver NLopt\footnote{https://nlopt.readthedocs.io/en/latest/}.
Cubic B-spline ($ p_b=3 $) is used as the trajectory representation.

To achieve fast exploration, an efficient mapping framework is essential.
In our work, we utilize a volumetric map \cite{han2019fiesta}, which has been successfully applied to fast autonomous flights\cite{zhou2019robust,zhou2020raptor} in complex scenes.
Similar to \cite{wurm2010octomap}, which is widely applied in exploration, \cite{han2019fiesta} builds an occupancy grid representation of the space.
Meanwhile it also maintains an ESDF incrementally to facilitate the trajectory planning.
For brevity we refer interested readers to \cite{han2019fiesta} for more details about our mapping framework.

In all field experiments, we localize the quadrotor by a visual-inertial state estimator\cite{qin2018vins}.
We use a geometric controller\cite{lee2010} for tracking control of the $ (x,y,z,\xi) $ trajectory.
We equipped our customized quadrotor platform with an Intel RealSense Depth Camera D435i.
All the above modules run on an Intel Core i7-8550U CPU.

\begin{figure}[t]
	\begin{center}          
		{\includegraphics[width=0.8\columnwidth]{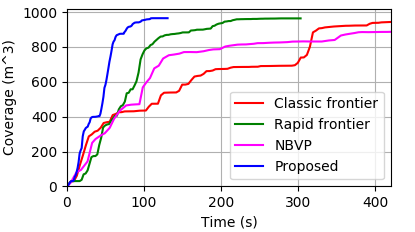}}       
		{\includegraphics[width=0.8\columnwidth]{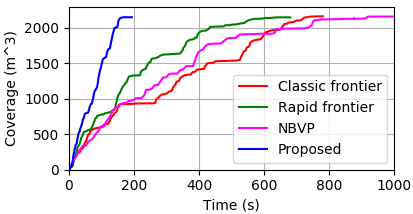}}       
      \vspace{-0.5cm}
	\end{center}
   \caption{\label{fig:progress} The exploration progress of four methods in the bridge (top) and large maze (bottom) scenarios.
   }
   \vspace{-0.3cm}
\end{figure} 

\subsection{Benchmark and Analysis}
\label{subs:results_benchmark}

\begin{table}
   \centering
   \caption{\label{tab:computation_time} Average computation time of each proposed component.}
   \begin{tabular}{>{\centering}p{0.7cm}>{\centering}p{0.9cm}>{\centering}p{1.1cm}>{\centering}p{0.9cm}>{\centering}p{0.9cm}>{\centering}p{0.9cm}c} 
   \hline\hline
   \textbf{Scene}  & \multicolumn{6}{c}{\textbf{Average computation time (ms)}}                                                                                                                                                                                                                                                                                                                                          \\ 
   \cline{2-7}
                   & \multirow{2}{*}{\begin{tabular}[c]{@{}c@{}}Frontier\\Sect.\ref{subs:frontier_detection}\end{tabular}} & \multirow{2}{*}{\begin{tabular}[c]{@{}c@{}}View.+Cost\\Sect.\ref{subs:frontier_viewpoint}\end{tabular}} & \multirow{2}{*}{\begin{tabular}[c]{@{}c@{}}Global\\Sect.\ref{subs:planning_global}\end{tabular}} & \multirow{2}{*}{\begin{tabular}[c]{@{}c@{}}Local\\Sect.\ref{subs:planning_local}\end{tabular}} & \multirow{2}{*}{\begin{tabular}[c]{@{}c@{}}Traj.\\Sect.\ref{subs:planning_traj}\end{tabular}} & \multirow{2}{*}{Total}  \\
                   &                                                                          &                                                                           &                                                                      &                                                                      &                                                                      &                         \\
   Bridge          & 4.69                                                                     & 4.86+5.16                                                                 & 1.12                                                                 & 4.10                                                                 & 4.23                                                                 & 24.17                   \\
   Maze            & 5.21                                                                     & 6.06+10.97                                                                & 3.53                                                                 & 4.98                                                                 & 5.47                                                                 & 36.23                   \\
   \hline\hline
   \end{tabular}
   \vspace{-1.8cm}
\end{table}

We test our proposed framework in simulation.
We benchmark it in a bridge scenario and a large maze scenario. 
Three methods are compared: the NBVP \cite{bircher2016receding}, the classic frontier method\cite{yamauchi1997frontier} and the rapid frontier method\cite{cieslewski2017rapid}. 
Note that no open source code is available for \cite{cieslewski2017rapid}, so we use our implementation.
In all tests the dynamic limits are set as $ v_{\text{max}}=2.0$ m/s and $ \dot{\xi_{\text{max}}} = 0.9 $ rad/s for all methods.
The FOVs of the sensors are set as $ [80\times60] \ \text{deg} $ with a maximum range of $4.5 $ m.
In both scenarios each method is run for 3 times with the same initial configuration.
Statistics and exploration progresses of the four methods are shown in Tab.\ref{tab:benchmark} and \ref{fig:progress} respectively.
The computation time of each component of our method is listed in Tab.\ref{tab:computation_time}.

\subsubsection{Bridge Scenario}
\label{subsubs:bridge}

Firstly, we compare the four methods in a $ 10 \times 20 \times 5 \ \text{m}^3 $ space containing a bridge, as shown in Fig.\ref{fig:bridge_compare}.
The result indicates that we achieve much shorter exploration time and smaller time variance.
The overall exploration path of our method is significantly shorter, primarily because we plan tours globally. 
The executed path is smoother, since we refine motions locally and generate smooth trajectories.
Also, we are able to navigate at a higher flight speed, owing to the minimum-time trajectory planning.

\begin{figure}[t]
	\begin{center}          
		{\includegraphics[width=0.75\columnwidth]{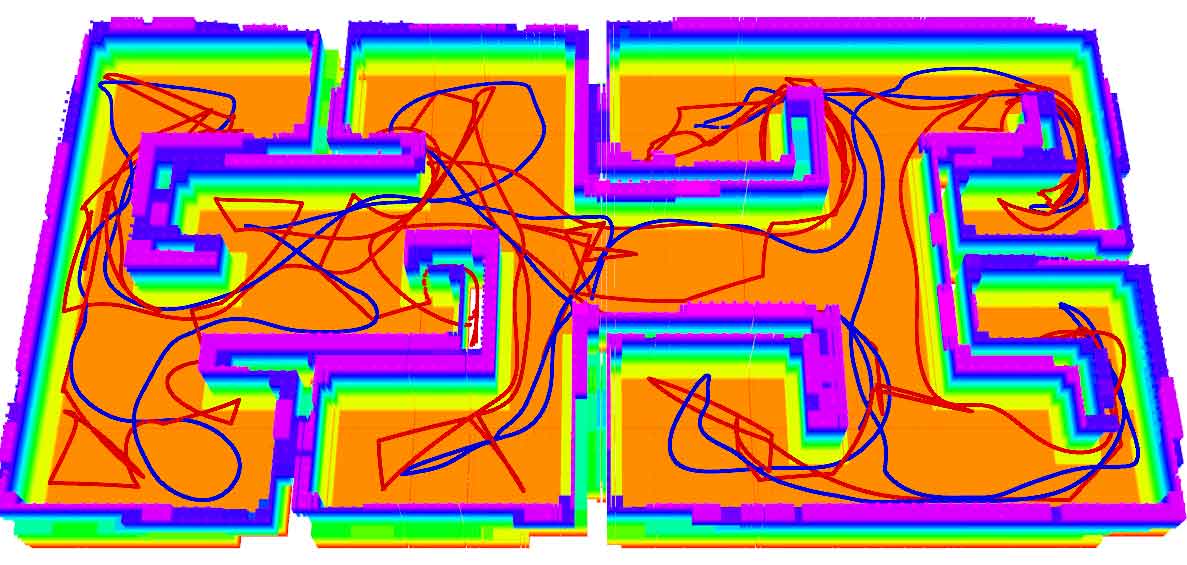}}       
		{\includegraphics[width=0.77\columnwidth]{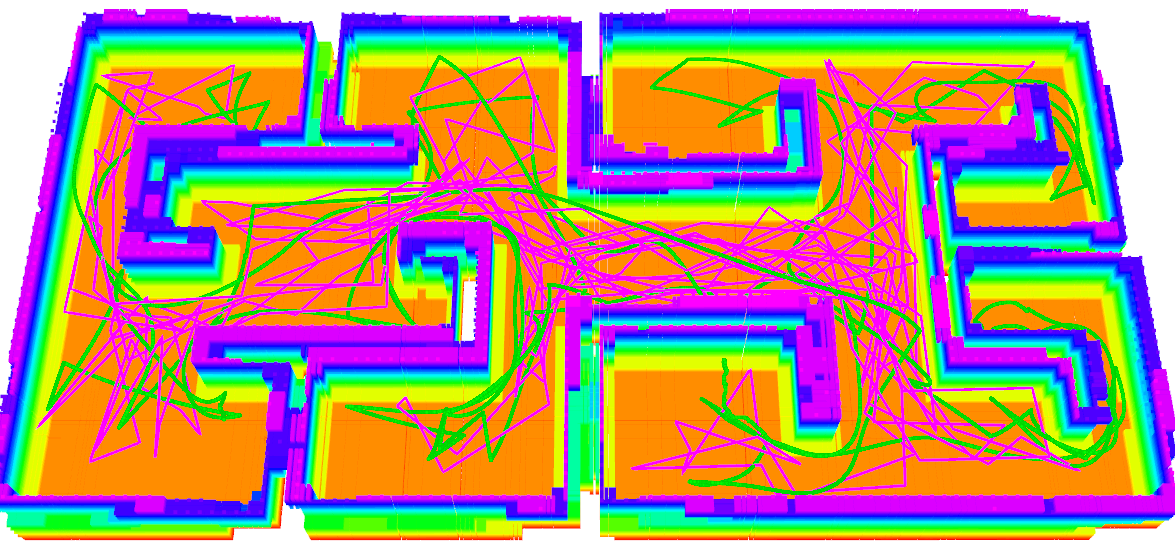}}       
      \vspace{-0.5cm}
	\end{center}
   \caption{\label{fig:four_compare_maze} Path generated by the proposed method (blue), classic frontier method\cite{yamauchi1997frontier}(red), rapid frontier method\cite{cieslewski2017rapid} (green) and NBVP\cite{bircher2016receding}(pink).
   }
   \vspace{-0.25cm}
\end{figure} 


\subsubsection{Large Maze Scenario}
\label{subsubs:maze}

We also compare the methods quantitatively in a large maze environment shown in Fig.\ref{fig:four_compare_maze}.
The explored space is $ 20 \times 80 \times 3 \ \text{m}^3 $ large.
In this scenario, all the benchmarked methods take a long time to reach full coverage, due to the complexity of the scene.
In contrast, our method completes the exploration 4+ times faster on average. 
Path executed by the four methods after completion are displayed in Fig.\ref{fig:four_compare_maze}.
Noticeably, our method explores the maze in a more sensible order, without revisiting the same place frequently.
In consequence, it produces a much shorter coverage path and an approximately linear exploration rate (Fig.\ref{fig:progress}).
This behavior is owing to the global plan, without which known regions may be revisited many times and slow down the progress, as the benchmarked methods do. 
Also note that the computation time in the large maze is longer, mostly due to the larger scene, which naturally involves a greater number of frontier clusters.



\begin{figure}[t]
	\begin{center}          
		{\includegraphics[width=0.493\columnwidth]{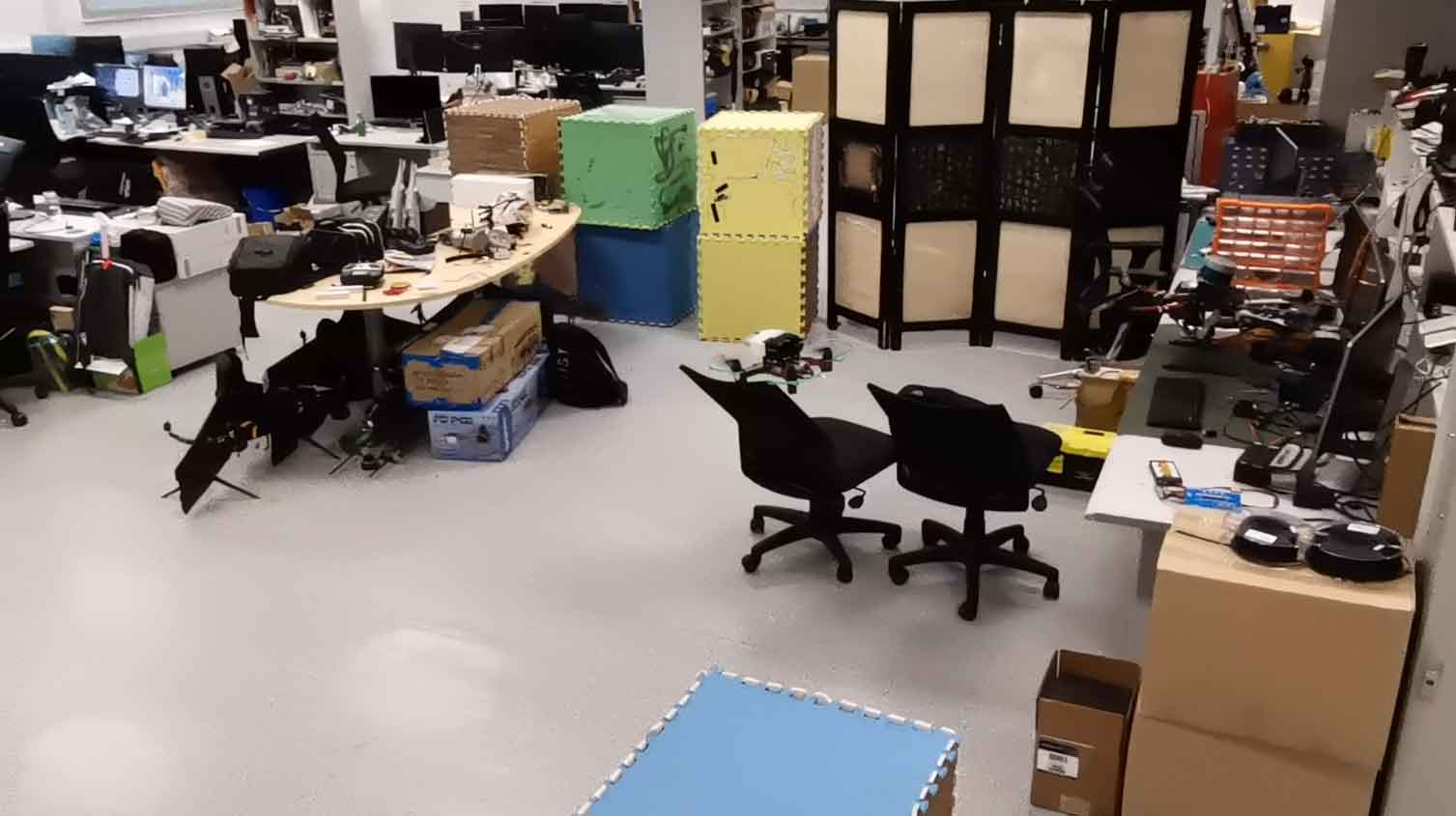}}       
		{\includegraphics[width=0.493\columnwidth]{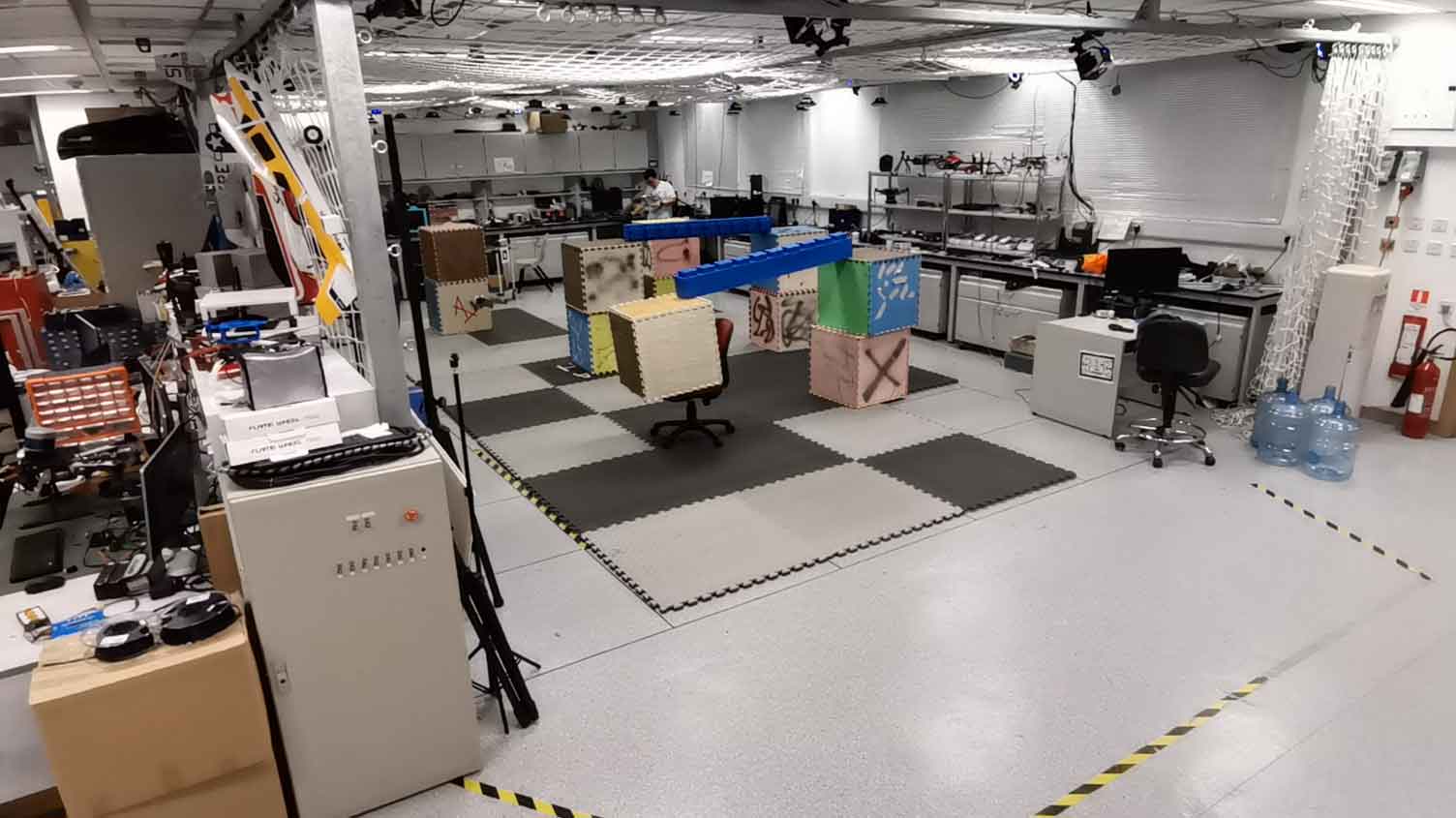}}       
		{\includegraphics[width=0.75\columnwidth]{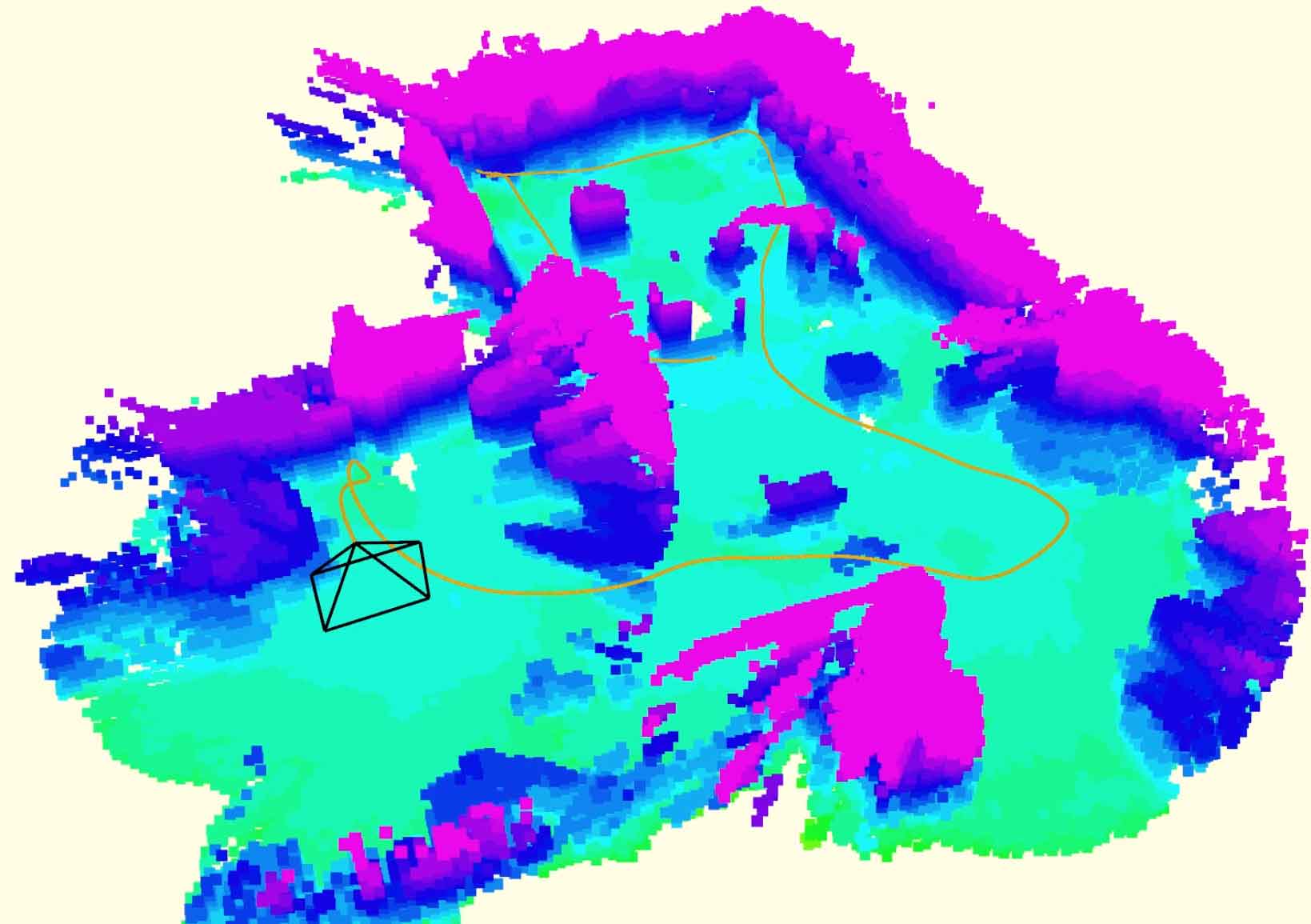}}       
    \vspace{-0.5cm}
	\end{center}
  \caption{\label{fig:indoor} Experiments in an indoor scene composed of two room: a small one with tables and chairs (top left), a large room cluttered with obstacles (top right).
  }
  \vspace{-1.4cm}
\end{figure} 

\begin{figure}[t]
	\begin{center}          
		{\includegraphics[width=0.75\columnwidth]{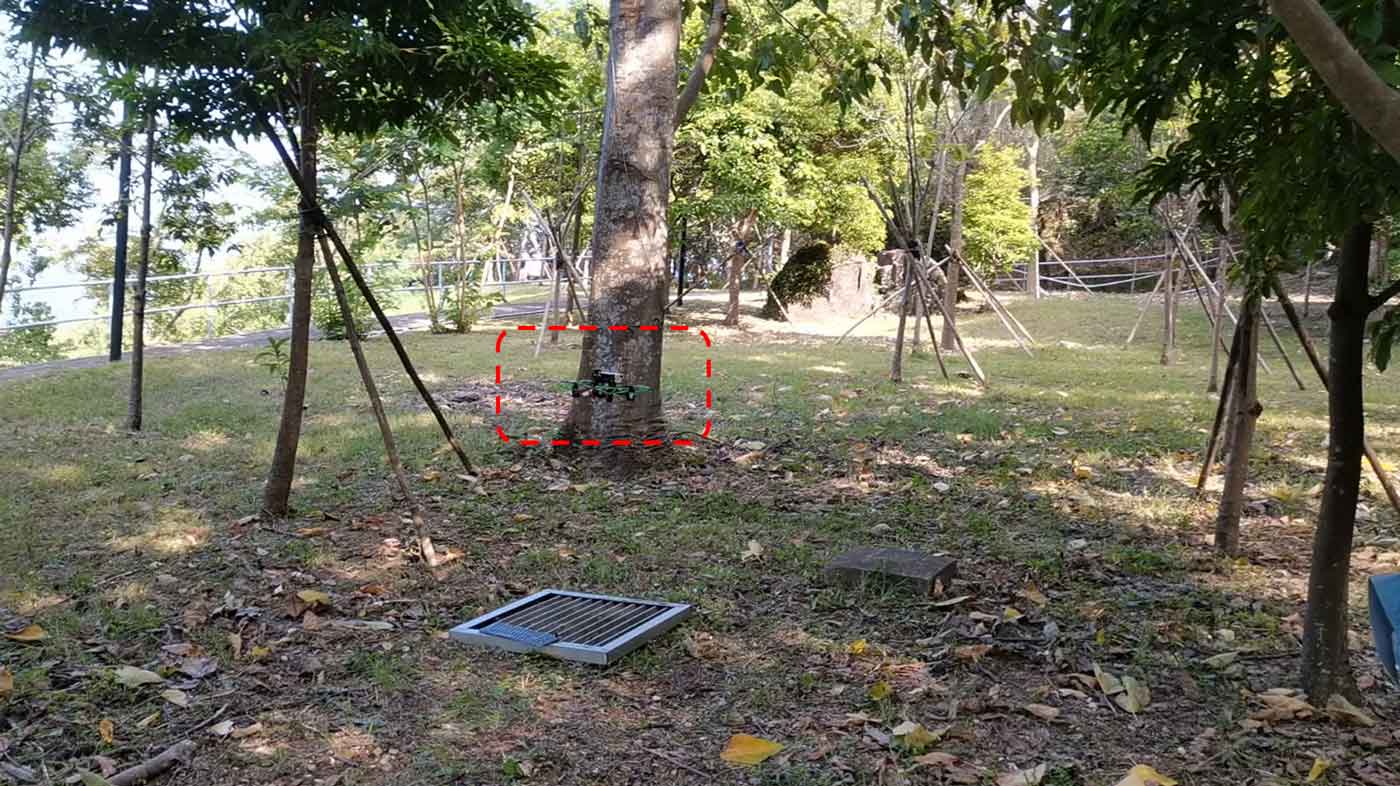}}       
		{\includegraphics[width=0.75\columnwidth]{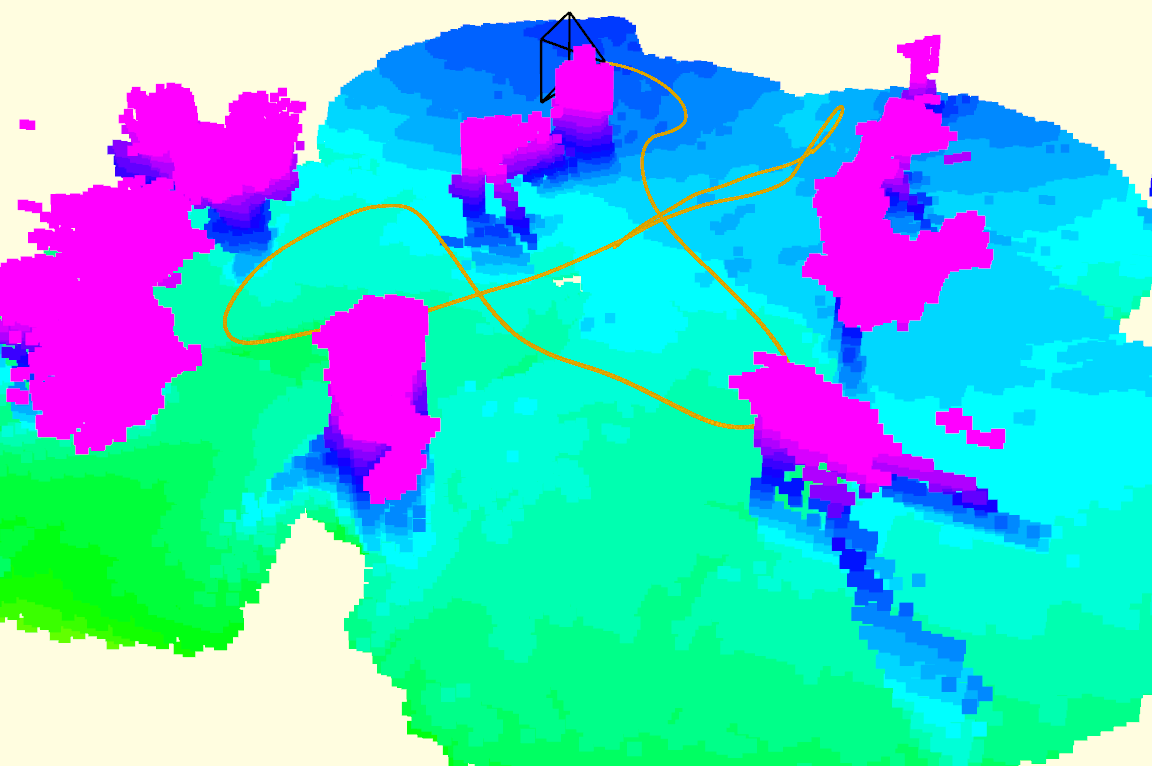}}       
      \vspace{-0.3cm}
	\end{center}
  \caption{\label{fig:forest} Exploration experiment conducted in a forest.
  }
  \vspace{-0.4cm}
\end{figure} 

\subsection{Field Exploration Tests}
\label{subs:results_field}

To further validate the proposed method, we conduct extensive field experiments in both indoor and outdoor environments.
In all tests we set the dynamics limits as $ v_{\text{max}}=1.5$ m/s, $ a_{\text{max}}=0.8 $ m/s and $ \dot{\xi}_{\text{max}} = 0.9 $ rad/s.
Note that we do not use any external device for localization and only rely on the onboard state estimator.

First, we present fast exploration tests in two indoor scenes.
The first scene is shown in Fig.\ref{fig:intro}, within which we deploy dozens of obstacles and the quadrotor should perform 3D maneuvers to map the unknown space and avoid obstacles simultaneously.
We bound the space to be explored with a $ 10 \times 6 \times 2  \ \text{m}^3 $ box.
One sample map and the flight trajectory is presented in Fig.\ref{fig:intro}.
The second indoor scene is a larger environment including two rooms, where one room is similar to scene 1 and the other is a part of an office containing tables and chairs.
The space is bounded by a $ 15 \times 11 \times 2 \ \text{m}^3 $ box.
The quadrotor starts by exploring the large room, after which it proceeds to the small one.
The second scene, the online generated map and trajectory are shown in Fig.\ref{fig:indoor}. 
Note that in the two scenes the quadrotor starts out at a spot with low visibility, so it only maps a small region of the environment at the beginning. 
Finally, to validate our method in natural environments, we conduct exploration tests in a forest. 
The size of the area to explore is $ 11 \times 10 \times 2 \ \text{m}^3 $.
The experiment environment and the associated results are displayed Fig.\ref{fig:forest}.

The above experiments demonstrate the capability of our method in complex real-world scenarios.
They also show the merits of our autonomous quadrotor system, among which the state estimation\cite{qin2018vins} and mapping modules\cite{han2019fiesta} are also crucial to fullfil the real-world tasks.
Video demonstration of all experiments is available (Fig.\ref{fig:intro}), we refer the readers to it for more details.

\vspace{-0.2cm}
\section{Conclusions}
\label{sec:conclude}
In this paper, we propose a hierarchical framework for rapid autonomous quadrotor exploration.
An incrementally maintained FIS is introduced to provide the exploration planning with essential information.
Based on FIS, a hierarchical planner plans exploration motions in three sequential steps, which finds efficient global tours, selects a local set of optimal viewpoints, and generates minimum-time local trajectories.
The method makes decisions at high frequency to respond quickly to environmental changes.
Both benchmark and real-world tests show the competence of our method.

One limitation of our method is assuming perfect state estimation, as most methods do.
We evaluate our method in simulation with ground truth localization, while drifts in pose are not considered.
However, error in state estimation exists generally and can not be ignored.
In the future we plan to consider the state estimation uncertainty in our method and evaluate its performance under pose drifts.

\addtolength{\textheight}{0.cm}   

\newlength{\bibitemsep}\setlength{\bibitemsep}{0.0\baselineskip}
\newlength{\bibparskip}\setlength{\bibparskip}{0.0pt}
\let\oldthebibliography\thebibliography
\renewcommand\thebibliography[1]{%
\oldthebibliography{#1}%
\setlength{\parskip}{\bibitemsep}%
\setlength{\itemsep}{\bibparskip}%
}

\bibliography{zby} 

\begin{thebibliography}{10}
\providecommand{\url}[1]{#1}
\csname url@samestyle\endcsname
\providecommand{\newblock}{\relax}
\providecommand{\bibinfo}[2]{#2}
\providecommand{\BIBentrySTDinterwordspacing}{\spaceskip=0pt\relax}
\providecommand{\BIBentryALTinterwordstretchfactor}{4}
\providecommand{\BIBentryALTinterwordspacing}{\spaceskip=\fontdimen2\font plus
\BIBentryALTinterwordstretchfactor\fontdimen3\font minus
  \fontdimen4\font\relax}
\providecommand{\BIBforeignlanguage}[2]{{%
\expandafter\ifx\csname l@#1\endcsname\relax
\typeout{** WARNING: IEEEtran.bst: No hyphenation pattern has been}%
\typeout{** loaded for the language `#1'. Using the pattern for}%
\typeout{** the default language instead.}%
\else
\language=\csname l@#1\endcsname
\fi
#2}}
\providecommand{\BIBdecl}{\relax}
\BIBdecl

\bibitem{cieslewski2017rapid}
T.~Cieslewski, E.~Kaufmann, and D.~Scaramuzza, ``Rapid exploration with
  multi-rotors: A frontier selection method for high speed flight,'' in
  \emph{Proc. of the {IEEE/RSJ} Intl. Conf. on Intell. Robots and
  Syst.({IROS})}.\hskip 1em plus 0.5em minus 0.4em\relax IEEE, 2017, pp.
  2135--2142.

\bibitem{schmid2020efficient}
L.~Schmid, M.~Pantic, R.~Khanna, L.~Ott, R.~Siegwart, and J.~Nieto, ``An
  efficient sampling-based method for online informative path planning in
  unknown environments,'' \emph{IEEE Robotics and Automation Letters}, vol.~5,
  no.~2, pp. 1500--1507, 2020.

\bibitem{meng2017two}
Z.~Meng, H.~Qin, Z.~Chen, X.~Chen, H.~Sun, F.~Lin, and M.~H. Ang~Jr, ``A
  two-stage optimized next-view planning framework for 3-d unknown environment
  exploration, and structural reconstruction,'' \emph{IEEE Robotics and
  Automation Letters}, vol.~2, no.~3, pp. 1680--1687, 2017.

\bibitem{selin2019efficient}
M.~Selin, M.~Tiger, D.~Duberg, F.~Heintz, and P.~Jensfelt, ``Efficient
  autonomous exploration planning of large-scale 3-d environments,'' \emph{IEEE
  Robotics and Automation Letters}, vol.~4, no.~2, pp. 1699--1706, 2019.

\bibitem{dharmadhikari2020motion}
M.~Dharmadhikari, T.~Dang, L.~Solanka, J.~Loje, H.~Nguyen, N.~Khedekar, and
  K.~Alexis, ``Motion primitives-based path planning for fast and agile
  exploration using aerial robots,'' in \emph{Proc. of the {IEEE} Intl. Conf.
  on Robot. and Autom. ({ICRA})}.\hskip 1em plus 0.5em minus 0.4em\relax IEEE,
  2020, pp. 179--185.

\bibitem{song2017online}
S.~Song and S.~Jo, ``Online inspection path planning for autonomous 3d modeling
  using a micro-aerial vehicle.'' in \emph{Proc. of the {IEEE} Intl. Conf. on
  Robot. and Autom. ({ICRA})}, 2017, pp. 6217--6224.

\bibitem{yamauchi1997frontier}
B.~Yamauchi, ``A frontier-based approach for autonomous exploration,'' in
  \emph{Proceedings 1997 IEEE International Symposium on Computational
  Intelligence in Robotics and Automation CIRA'97.'Towards New Computational
  Principles for Robotics and Automation'}.\hskip 1em plus 0.5em minus
  0.4em\relax IEEE, 1997, pp. 146--151.

\bibitem{julia2012comparison}
M.~Juli{\'a}, A.~Gil, and O.~Reinoso, ``A comparison of path planning
  strategies for autonomous exploration and mapping of unknown environments,''
  \emph{Auton. Robots}, vol.~33, no.~4, pp. 427--444, 2012.

\bibitem{shen2012stochastic}
S.~Shen, N.~Michael, and V.~Kumar, ``Stochastic differential equation-based
  exploration algorithm for autonomous indoor 3d exploration with a
  micro-aerial vehicle,'' \emph{Intl. J. Robot. Research ({IJRR})}, vol.~31,
  no.~12, pp. 1431--1444, 2012.

\bibitem{deng2020robotic}
D.~Deng, R.~Duan, J.~Liu, K.~Sheng, and K.~Shimada, ``Robotic exploration of
  unknown 2d environment using a frontier-based automatic-differentiable
  information gain measure,'' in \emph{2020 IEEE/ASME International Conference
  on Advanced Intelligent Mechatronics (AIM)}.\hskip 1em plus 0.5em minus
  0.4em\relax IEEE, 2020, pp. 1497--1503.

\bibitem{connolly1985determination}
C.~Connolly, ``The determination of next best views,'' in \emph{Proc. of the
  {IEEE} Intl. Conf. on Robot. and Autom. ({ICRA})}, vol.~2.\hskip 1em plus
  0.5em minus 0.4em\relax IEEE, 1985, pp. 432--435.

\bibitem{bircher2016receding}
A.~Bircher, M.~Kamel, K.~Alexis, H.~Oleynikova, and R.~Siegwart, ``Receding
  horizon" next-best-view" planner for 3d exploration,'' in \emph{Proc. of the
  {IEEE} Intl. Conf. on Robot. and Autom. ({ICRA})}.\hskip 1em plus 0.5em minus
  0.4em\relax IEEE, 2016, pp. 1462--1468.

\bibitem{papachristos2017uncertainty}
C.~Papachristos, S.~Khattak, and K.~Alexis, ``Uncertainty-aware receding
  horizon exploration and mapping using aerial robots,'' in \emph{2017 IEEE
  international conference on robotics and automation (ICRA)}.\hskip 1em plus
  0.5em minus 0.4em\relax IEEE, 2017, pp. 4568--4575.

\bibitem{dang2018visual}
T.~Dang, C.~Papachristos, and K.~Alexis, ``Visual saliency-aware receding
  horizon autonomous exploration with application to aerial robotics,'' in
  \emph{2018 IEEE International Conference on Robotics and Automation
  (ICRA)}.\hskip 1em plus 0.5em minus 0.4em\relax IEEE, 2018, pp. 2526--2533.

\bibitem{bircher2018receding}
A.~Bircher, M.~Kamel, K.~Alexis, H.~Oleynikova, and R.~Siegwart, ``Receding
  horizon path planning for 3d exploration and surface inspection,''
  \emph{Auton. Robots}, vol.~42, no.~2, pp. 291--306, 2018.

\bibitem{witting2018history}
C.~Witting, M.~Fehr, R.~B{\"a}hnemann, H.~Oleynikova, and R.~Siegwart,
  ``History-aware autonomous exploration in confined environments using mavs,''
  in \emph{2018 IEEE/RSJ International Conference on Intelligent Robots and
  Systems (IROS)}.\hskip 1em plus 0.5em minus 0.4em\relax IEEE, 2018, pp. 1--9.

\bibitem{wang2019efficient}
C.~Wang, D.~Zhu, T.~Li, M.~Q.-H. Meng, and C.~W. de~Silva, ``Efficient
  autonomous robotic exploration with semantic road map in indoor
  environments,'' \emph{IEEE Robotics and Automation Letters}, vol.~4, no.~3,
  pp. 2989--2996, 2019.

\bibitem{charrow2015information}
B.~Charrow, G.~Kahn, S.~Patil, S.~Liu, K.~Goldberg, P.~Abbeel, N.~Michael, and
  V.~Kumar, ``Information-theoretic planning with trajectory optimization for
  dense 3d mapping.'' in \emph{Proc. of Robot.: Sci. and Syst. ({RSS})},
  vol.~11, 2015.

\bibitem{MelKum1105}
D.~Mellinger and V.~Kumar, ``Minimum snap trajectory generation and control for
  quadrotors,'' in \emph{Proc. of the {IEEE} Intl. Conf. on Robot. and Autom.
  ({ICRA})}, Shanghai, China, May 2011, pp. 2520--2525.

\bibitem{RicBryRoy1312}
C.~Richter, A.~Bry, and N.~Roy, ``Polynomial trajectory planning for aggressive
  quadrotor flight in dense indoor environments,'' in \emph{Proc. of the Intl.
  Sym. of Robot. Research ({ISRR})}, Dec. 2013, pp. 649--666.

\bibitem{CheLiuShe2016}
J.~Chen, T.~Liu, and S.~Shen, ``Online generation of collision-free
  trajectories for quadrotor flight in unknown cluttered environments,'' in
  \emph{Proc. of the {IEEE} Intl. Conf. on Robot. and Autom. ({ICRA})},
  Stockholm, Sweden, May 2016, pp. 1476--1483.

\bibitem{fei2018icra}
F.~Gao, W.~Wu, Y.~Lin, and S.~Shen, ``Online safe trajectory generation for
  quadrotors using fast marching method and bernstein basis polynomial,'' in
  \emph{Proc. of the {IEEE} Intl. Conf. on Robot. and Autom. ({ICRA})},
  Brisbane, Australia, May 2018.

\bibitem{ding2019efficient}
W.~Ding, W.~Gao, K.~Wang, and S.~Shen, ``An efficient b-spline-based
  kinodynamic replanning framework for quadrotors,'' \emph{IEEE Transactions on
  Robotics}, vol.~35, no.~6, pp. 1287--1306, 2019.

\bibitem{tordesillas2019faster}
J.~Tordesillas, B.~T. Lopez, and J.~P. How, ``{FASTER}: Fast and safe
  trajectory planner for flights in unknown environments,'' in \emph{Proc. of
  the {IEEE/RSJ} Intl. Conf. on Intell. Robots and Syst.({IROS})}.\hskip 1em
  plus 0.5em minus 0.4em\relax IEEE, 2019.

\bibitem{oleynikova2016continuous}
H.~Oleynikova, M.~Burri, Z.~Taylor, J.~Nieto, R.~Siegwart, and E.~Galceran,
  ``Continuous-time trajectory optimization for online uav replanning,'' in
  \emph{Proc. of the {IEEE/RSJ} Intl. Conf. on Intell. Robots and
  Syst.({IROS})}, Daejeon, Korea, Oct. 2016, pp. 5332--5339.

\bibitem{fei2017iros}
F.~Gao, Y.~Lin, and S.~Shen, ``Gradient-based online safe trajectory generation
  for quadrotor flight in complex environments,'' in \emph{Proc. of the
  {IEEE/RSJ} Intl. Conf. on Intell. Robots and Syst.({IROS})}, Sept 2017, pp.
  3681--3688.

\bibitem{usenko2017real}
V.~Usenko, L.~von Stumberg, A.~Pangercic, and D.~Cremers, ``Real-time
  trajectory replanning for mavs using uniform b-splines and a 3d circular
  buffer,'' in \emph{Proc. of the {IEEE/RSJ} Intl. Conf. on Intell. Robots and
  Syst.({IROS})}.\hskip 1em plus 0.5em minus 0.4em\relax IEEE, 2017, pp.
  215--222.

\bibitem{zhou2019robust}
B.~Zhou, F.~Gao, L.~Wang, C.~Liu, and S.~Shen, ``Robust and efficient quadrotor
  trajectory generation for fast autonomous flight,'' \emph{IEEE Robotics and
  Automation Letters}, vol.~4, no.~4, pp. 3529--3536, 2019.

\bibitem{zhou2020robust}
B.~Zhou, F.~Gao, J.~Pan, and S.~Shen, ``Robust real-time uav replanning using
  guided gradient-based optimization and topological paths,'' in \emph{Proc. of
  the {IEEE} Intl. Conf. on Robot. and Autom. ({ICRA})}.\hskip 1em plus 0.5em
  minus 0.4em\relax IEEE, 2020, pp. 1208--1214.

\bibitem{zhou2020raptor}
B.~Zhou, J.~Pan, F.~Gao, and S.~Shen, ``Raptor: Robust and perception-aware
  trajectory replanning for quadrotor fast flight,'' \emph{arXiv preprint
  arXiv:2007.03465}, 2020.

\bibitem{ratliff2009chomp}
N.~Ratliff, M.~Zucker, J.~A. Bagnell, and S.~Srinivasa, ``Chomp: Gradient
  optimization techniques for efficient motion planning,'' in \emph{Proc. of
  the {IEEE} Intl. Conf. on Robot. and Autom. ({ICRA})}, May 2009, pp.
  489--494.

\bibitem{ericson2004real}
C.~Ericson, \emph{Real-time collision detection}.\hskip 1em plus 0.5em minus
  0.4em\relax CRC Press, 2004.

\bibitem{helsgaun2000effective}
K.~Helsgaun, ``An effective implementation of the lin--kernighan traveling
  salesman heuristic,'' \emph{European Journal of Operational Research}, vol.
  126, no.~1, pp. 106--130, 2000.

\bibitem{han2019fiesta}
L.~Han, F.~Gao, B.~Zhou, and S.~Shen, ``Fiesta: Fast incremental euclidean
  distance fields for online motion planning of aerial robots,'' \emph{arXiv
  preprint arXiv:1903.02144}, 2019.

\bibitem{wurm2010octomap}
K.~M. Wurm, A.~Hornung, M.~Bennewitz, C.~Stachniss, and W.~Burgard, ``Octomap:
  A probabilistic, flexible, and compact 3d map representation for robotic
  systems,'' in \emph{Proc. of the {IEEE} Intl. Conf. on Robot. and Autom.
  ({ICRA})}, vol.~2, Anchorage, AK, US, May 2010.

\bibitem{qin2018vins}
T.~Qin, P.~Li, and S.~Shen, ``Vins-mono: A robust and versatile monocular
  visual-inertial state estimator,'' \emph{{IEEE} Trans. Robot. ({TRO})},
  vol.~34, no.~4, pp. 1004--1020, 2018.

\bibitem{lee2010}
T.~Lee, M.~Leoky, and N.~H. McClamroch, ``Geometric tracking control of a
  quadrotor uav on se (3),'' in \emph{Proc. of the {IEEE} Control and Decision
  Conf. ({CDC})}, Atlanta, GA, Dec. 2010, pp. 5420--5425.

\end{thebibliography}

\end{document}